\documentclass[letterpaper]{article} 
\PassOptionsToPackage{table}{xcolor} 
\usepackage{aaai25}  
\usepackage{times}  
\usepackage{helvet}  
\usepackage{courier}  
\usepackage[hyphens]{url}  
\usepackage{graphicx} 
\usepackage{natbib}  
\usepackage{caption} 
\usepackage{algorithm}
\usepackage{algorithmic}
\usepackage{pifont}

\usepackage{newfloat}
\usepackage{listings}
\DeclareCaptionStyle{ruled}{labelfont=normalfont,labelsep=colon,strut=off} 
\floatstyle{ruled}
\newfloat{listing}{tb}{lst}{}
\floatname{listing}{Listing}
\title{Multi-Modal Recommendation Unlearning\\ for Legal, Licensing, and Modality Constraints}
\author{
    Yash Sinha\textsuperscript{\rm 1},
    Murari Mandal\textsuperscript{\rm 2},
    Mohan Kankanhalli\textsuperscript{\rm 1}
}
\affiliations{
    \textsuperscript{\rm 1}School of Computing, National University of Singapore\\
    \textsuperscript{\rm 2}RespAI Lab, School of Computer Engineering, KIIT Bhubaneswar, India\\
    yashsinha@comp.nus.edu.sg, murari.mandalfcs@kiit.ac.in, mohan@comp.nus.edu.sg
}

\usepackage{xspace} 
\usepackage{amsfonts}
\newcommand{\name}{\textsc{MMRecUn}\xspace}
\newcommand{\G}{\mathcal{G}\xspace}
\newcommand{\V}{\mathcal{V}\xspace}
\newcommand{\E}{\mathcal{E}\xspace}
\newcommand{\D}{\mathcal{D}\xspace}

\usepackage{tikz}

\usepackage{booktabs}
\usepackage{multirow}
\usepackage{graphicx}
\usepackage{amsmath}
\usepackage{lscape}
\usepackage{float} 

\newtheorem{prop}{Proposition}
\usepackage[table]{xcolor}
\usepackage{subcaption}
\newcolumntype{H}{>{\setbox0=\hbox\bgroup}c<{\egroup}@{}} 

\begin{document}
\maketitle


\begin{abstract}
User data spread across multiple modalities has popularized multi-modal recommender systems (MMRS). They recommend diverse content such as products, social media posts, TikTok reels, etc., based on a user-item interaction graph. With rising data privacy demands, recent methods propose unlearning private user data from uni-modal recommender systems (RS). However, methods for unlearning item data related to outdated user preferences, revoked licenses, and legally requested removals are still largely unexplored.

Previous RS unlearning methods are unsuitable for MMRS due to the incompatibility of their matrix-based representation with the multi-modal user-item interaction graph. Moreover, their data partitioning step degrades performance on each shard due to poor data heterogeneity and requires costly performance aggregation across shards.

This paper introduces MMRecUn, the first approach known to us for unlearning in MMRS and unlearning item data. Given a trained RS model, MMRecUn employs a novel Reverse Bayesian Personalized Ranking (BPR) objective to enable the model to forget marked data. The reverse BPR attenuates the impact of user-item interactions within the forget set, while the forward BPR reinforces the significance of user-item interactions within the retain set.
Our experiments demonstrate that MMRecUn outperforms baseline methods across various unlearning requests when evaluated on benchmark MMRS datasets. MMRecUn achieves recall performance improvements of up to 49.85\% compared to baseline methods and is up to 1.3× faster than the Gold model, which is trained on retain set from scratch. MMRecUn offers significant advantages, including superiority in removing target interactions, preserving retained interactions, and zero overhead costs compared to previous methods.

\end{abstract}

%
\begin{links}
    \link{Code}{https://github.com/MachineUnlearn/MMRecUN}
    \link{Extended version}{https://arxiv.org/abs/2405.15328}
\end{links}

\section{Introduction}
Recommender systems (RS)~\cite{smith2017two, He@LightGCN} leverage techniques like collaborative filtering, content-based filtering, and neural networks to recommend diverse content such as movies, music, products, news articles, and more, thereby boosting user engagement and satisfaction.
As user data increasingly spans multiple modalities~\cite{covington2016deep}, multi-modal recommender systems (MMRS) are gaining traction~\cite{wu2022survey, zhou2023comprehensive}. Recent advancements~\cite{yu2023multi, liu2023multimodal, wei2020graph, wei2019mmgcn, xu2018graphcar} have focused on capturing user preferences through both behavior data and diverse multi-modal item information. With growing concerns over data privacy, regulations like GDPR~\cite{voigt2017eu}
emphasize the importance of data protection and the ``right to be forgotten." While significant progress has been made in unlearning user data to enhance privacy~\cite{chen2022recommendation, li2023ultrare}, the unlearning of item data remains largely unexplored. This paper introduces \name, which, to the best of our knowledge, is the first attempt to address unlearning in MMRS and unlearning item data.

\textbf{Motivation.} Recent needs for dynamic recommendations.

\underline{\textit{Complex content licensing agreements}}: Universal Music Group’s (UMG) recent decision to pull its library from TikTok~\cite{Johnson_2024} silenced millions of user-created videos, necessitating TikTok's recommendations to adapt to the absence of UMG content. Similarly, the 2020 lawsuit~\cite{Deahl_2018} against Spotify by Wixen Music Publishing over songwriter compensation and the 2016 dispute~\cite{Levine_2018} between YouTube and Warner Music Group, which led to temporary content removal, highlight potential conflicts between existing models and evolving contractual obligations. These scenarios highlight the need for a dynamic RS that can adapt swiftly.

\underline{\textit{Legal compliance}}: The Algorithmic Accountability Act~\cite{Gursoy_Kennedy_Kakadiaris_2022} emphasizes the responsibility of companies to evaluate their machine learning systems for bias and discrimination. Current RS often reinforce user preferences, perpetuating stereotypes and creating filter bubbles and echo chambers~\cite{chen2023bias, lin2021mitigating}. This highlights the need for an RS that balances personalization with fairness and diversity.

\underline{\textit{Evolving user interests in different modalities}}: Imagine a user who usually posts about fitness and healthy living. Over time, she develops a new passion for travel photography, sharing photos and stories from her trips to exotic locations. Although her images and hashtags now reflect her love for travel, the RS still prioritizes fitness-related content. By analyzing both text and images, the system can adapt to her new interests, offering more relevant recommendations and enhancing her browsing experience.

\underline{\textit{Selective unlearning based on modalities}}: When Universal Music Group (UMG) removed its library from TikTok, the RS had to adapt to the absence of UMG audio content while still considering other modalities like videos. In MMRS, both the graph structure and feature embeddings are closely linked. Unlearning interactions in one modality, such as audio, can affect the entire RS, requiring careful adjustments to maintain accurate and effective recommendations.

\underline{\textit{Aiding selective transfer}}: Data privacy concerns and strict protection policies~\cite{liu2023differentially} challenge cross-domain RS~\cite{liureducing}. While sharing user-item data can be beneficial, it risks negative transfer, such as using horror movie ratings to recommend comedies. Unlearning can remove irrelevant data from the source domain before transfer, enhancing recommendation accuracy.

Dynamic MMRS that handle both user and item data can effectively tackle these multifaceted challenges. By enabling systems to dynamically update their models and remove outdated or irrelevant content, unlearning methods can significantly enhance user privacy, ensure legal compliance, adapt to changing content licenses, and evolve with user preferences. These systems can also combat recommendation bias and data poisoning. Moreover, they can reduce the GPU-hours carbon footprint by minimizing the need for repeated, resource-intensive retraining from scratch.\par

\textbf{Background and Related Work.} 
\textit{Given these benefits, one might ask: Can unlearning methods for uni-modal systems be adapted for the unique challenges of multi-modal systems?}
Previous unlearning methods~\cite{nguyen2022survey} face the following challenges:
\ding{182}~MMRS integrate diverse user-item data, including images, text, and behavior, into a unified convolutional graph. This is incompatible with the matrices, latent factors, feature representations, and temporal sequences used in matrix factorization~\cite{liu2023recommendation, xu2022netflix, liu2022forgetting, zhang2023closed}, collaborative filtering~\cite{li2024making, schelter2023forget}, and sequential RS~\cite{ye2023sequence}, respectively.
\ding{183}~For methods that do use graph-based representations, integrating diverse modalities is challenging. Because structure and feature embeddings are tightly integrated, unlearning in one modality has cascading effects on the other~\cite{cheng2023multimodal}.
\ding{184}~\textit{Approximate unlearning methods}~\cite{you2024rrl,zhang2023recommendation,li2023selective} involve expensive operations, such as inverting Hessian matrix or the Fisher Information Matrix. These computations are costly and impractical for MMRS having a large number of feature dimensions. 
\ding{185}~\textit{Exact unlearning methods} \cite{chen2022recommendation, li2023ultrare} adapt \textsc{SISA}~\cite{bourtoule2021machine} to split the dataset into shards. This disrupts the graph structure and degrades performance within each shard due to limited data and poor data heterogeneity~\cite{koch2023no} by $10-30\%$~\cite{you2024rrl}. The aggregation step, to preserve data structure and aggregate performance across shards, introduces significant overhead costs, impacting both training and inference, increasing proportionally with the number of shards~\cite{ramezani2021learn}. In worst case, when data that needs to be forgotten come from multiple shards, efficiency falls to the level of retraining from scratch.
\ding{186}~While unlearning might appear efficient by retraining only specific shards, the process of setting up and aggregating these shards introduces additional, unaccounted overhead. If the data is initially partitioned by user dimensions, it cannot be re-split for item unlearning. Consequently, simultaneous unlearning of both user and item data becomes impossible.

These constraints underscore the need for methods specifically designed for unlearning in MMRS and for unlearning item data. 
The challenges and previous works, including ~\cite{cheng2023multimodal, yuan2023federated, li2023making, chen2024post, ganhor2022unlearning, xin2024effectiveness, wang2024towards, sinha2023distill, tarun2023fast, chundawat2023zero, tarun2023deep, chundawat2023can}, provide valuable insights into this domain.

\textbf{Contributions.} They are summarized as follows:
\begin{enumerate}
    \item \name is the first approach known to us for unlearning in MMRS and unlearning item data.

    \item Our work addresses various unlearning requests, including single interaction removal, user preference adjustments, bias elimination, account deletion, and item removal, as shown in Table~\ref{tab:unlearningRequests}. 
    
    \item We define three properties for measuring unlearning in MMRS and introduce \textit{item-centric metrics} alongside traditional user-centric metrics. We also propose BPR divergence as a robust alternative to KL divergence for comparing recommendation scores.

    \item The experiments demonstrate that \name outperforms the baselines in various unlearning scenarios: user, item and user-item (simultaneous) unlearning. It achieves recall performance improvements of up to $\mathbf{49.85\%}$ compared to the baseline methods. It is up to $\mathbf{1.3}\times$ faster than the \textsc{Gold} model, which is trained on retain data from scratch. Moreover, \name offers enhanced efficiency, superior performance in removing target elements, preservation of performance for retained elements, and minimal overhead costs.
\end{enumerate}
\begin{table}[tb]
\centering
\resizebox{\columnwidth}{!}{%
\begin{tabular}{@{}p{1.7cm}p{0.6cm}p{3.1cm}p{5.3cm}@{}}
\toprule
\textbf{Level}           & \textbf{Qty} & \textbf{Catering to}             & \textbf{Example}                                                \\ \midrule
Interaction     & Single         & Privacy      & Remove a watched movie                \\
User Pref. & Many           & Evolving preferences    & Disinterest in Apple products          \\
Biased Item & Many & Bias elimination & Amazon tackling fake reviews  \\
Account         & All            & Privacy laws & User deletes account \\
License         & All            & Licensing agreements    & UMG removing library from TikTok              \\ \bottomrule
\end{tabular}%
}
\caption{Unlearning request types in MMRS, addressing privacy, preferences, bias elimination, and legal compliance.}
\label{tab:unlearningRequests}
\end{table}
\section{Preliminaries}
\label{sec:prelim}
\textbf{MMRS.} Let $\mathcal{U} = \{u\}$ denote the set of users and $\mathcal{I} = \{i\}$ denote the set of items. The embedding matrix for users is denoted as $\textbf{E}_{u} \in \mathbb{R}^{d \times |\mathcal{U}|}$, where $d$ is the embedding dimension. Similarly, the embedding matrices for each item modality are represented as $\textbf{E}_{i,m} \in \mathbb{R}^{d_m \times |\mathcal{I}|}$, where $d_m$ is the dimension of the features, $m \in \mathcal{M}$ denotes the modality, and $\mathcal{M} = \{v, t\}$, visual and textual, is the set of modalities considered. The historical behavior data of users is represented by matrix $\mathcal{Y} \in \{0, 1\}^{|\mathcal{U}| \times |\mathcal{I}|}$, where each entry $y_{u,i}$ indicates whether user $u$ interacted with item $i$. This data can be interpreted as a sparse behavior graph $\G = \{\V, \E\}$, where $\V = \{\mathcal{U} \cup \mathcal{I}\}$ denotes the set of nodes and $\E = \{(u,i) \mid u \in \mathcal{U}, i \in \mathcal{I}, y_{u,i} = 1\}$ denotes the set of edges. 
Let the model be denoted by $M(\cdot, \varphi)$ with parameters $\varphi$ which aims to encode collaborative signals latent in the interaction matrix $\mathcal{Y}$. 
The objective of MMRS is to accurately predict users' preferences by ranking items for each user based on predicted preference scores $\hat{y}_{u,i}$.

\textbf{Unlearning.} 
Let $\D =\{(u,i), {y}_{u,i}\}^{|\E|}, (u,i) \in \E$ represent a dataset of user-item interactions, split for training, validation and testing as $\D_T$, $\D_v$, and $\D_t$, respectively. The aim is to forget a set of data points, represented by $\D_f = \E_f \subseteq \E_T$, while retaining another set of data points, represented by $\D_r = \E_r$. It holds that $\D_r \bigcup \D_f = \D_T$ and $\D_r \bigcap \D_f = \emptyset$. If a node, i.e., a user or an item is marked for forgetting, then all interactions involving that user or item are also marked for forgetting. $(u, i) \in \E_f \implies \{(u', i') \mid u' = u \lor i' = i\} \subseteq \mathcal{E}_f$, where $(u, i)$ represents the interaction between user $u$ and item $i$, and $(u', i')$ represents any interaction involving the same user or item. Given an input $x$, the model's output is $M(x,\varphi)$. For a machine learning algorithm $A$, it generates model parameters as $\varphi = A(\D_T)$. A \textit{gold model} is trained from scratch only on the retain set $\D_r$, denoted by $\varphi_r = A(\D_r)$. An unlearning algorithm $U$ utilizes all or a subset of $\D_r$ and $\D_f$, as well as the original model $\varphi$ to generate an unlearned model $\varphi_u$. Hence, $\varphi_u = U(\varphi, \D_r, \D_f).$

\textbf{Problem Formulation.} Given a sparse behavior graph $\G$ and a model $M(\cdot, \varphi)$,  devise an unlearning algorithm $U$, that unlearns the forget set $\D_f$ to obtain an unlearned model $M(\cdot, \varphi_u)$ with updated parameters such that $\varphi_u$ closely approximates the performance of the gold model:
\begin{align}
    \mathcal{P}(M(x, \varphi_u) = y) \approx \mathcal{P}(M(x, \varphi_r) = y), \quad \forall x \in \D
\end{align}
where $\mathcal{P}(X)$ is the distribution of random variable $X$. 

\textbf{Properties.} Let 
$ \forall x \in \D$, $f(\D, \epsilon)$ be defined as,
    $f(\D, \epsilon): \mathcal{P}(M(x, \varphi_u)=y) - \mathcal{P}(M(x, \varphi_r)=y) \leq \epsilon. $
For close performance approximation, the unlearned model must possess:\par

\underline{\textit{Unlearning Specificity $f(D_f, \epsilon_f)$.}} The unlearning algorithm $U$ should effectively remove the influence of entities in $\D_f$, with $\varphi_u$ aligning closely with the gold model's probability distribution. 
A high $\epsilon_f$ indicates unsuccessful unlearning, while a low $\epsilon_f$ may signal a \textit{Streisand} effect, making the forget set more noticeable. Ideally, $\epsilon_f$ should approach zero.\par
\underline{\textit{Retention Fidelity $f(D_t, \epsilon_t)$.}} Equally important is preserving performance of the model. So the probability distribution on test set should align. 
A high value of $\epsilon_t$ indicates that the model's utility is compromised, as it becomes excessively tailored to the specific traits of the retain set. On the other hand, if $\epsilon_t$ is low, it indicates that the unlearning process has inadvertently led to the loss of characteristics of the retain set. So, $\epsilon_t$ should tend to zero.\par

\underline{\textit{Unlearning Generalizability $f(D_v, \epsilon_v)$.}} The unlearning algorithm $U$ must ensure that the process does not introduce biases or distortions that impair the model's generalization. This is evaluated by comparing scores on unseen data, confirming that unlearning preserves the model's performance on data excluded from training and unlearning.
A high $\epsilon_v$ indicates that unlearning may have overly tailored the model to the retain set, reducing generalization. Conversely, a low $\epsilon_v$ suggests that unlearning effectively retained the retain set's characteristics while removing the forget set's influence.\par



\underline{\textit{Unlearning Efficiency.}} The unlearning process should be faster than retraining the model from scratch. It should be efficient in terms of time and computational resources.\par


\begin{figure}[t]
    \centering
    \includegraphics[width=\columnwidth]
    {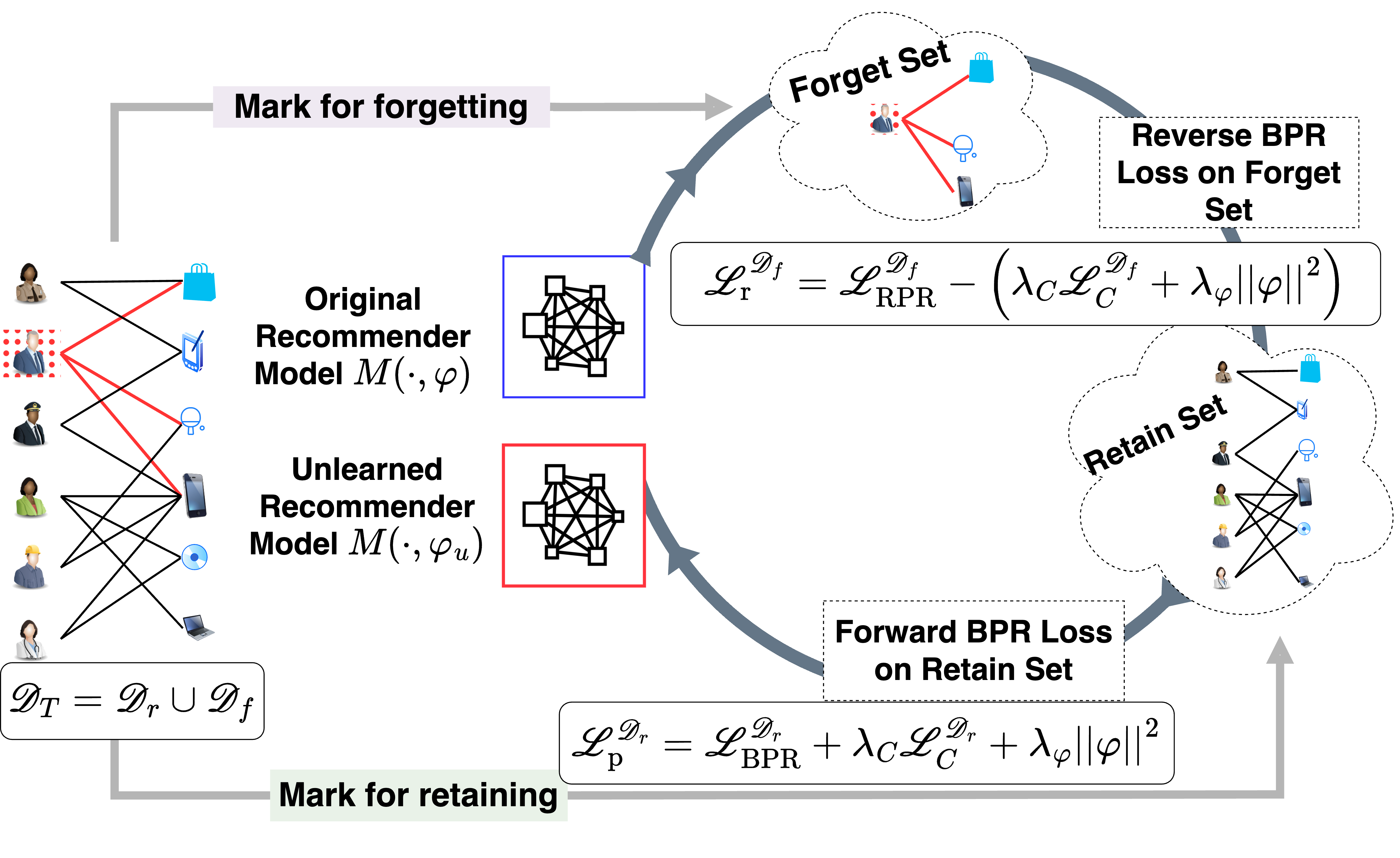}
    \caption{The proposed \name method illustrated. Adapts \textsc{Mgcn}'s architecture to unlearn multi-modal data while balancing retention fidelity and unlearning specificity.}
    \label{fig:reco_unlearning}
\end{figure}

\begin{table*}[t]
\centering
\resizebox{.88\textwidth}{!}{%
\begin{tabular}{@{}lllllllllllll@{}}
\toprule
 & \multicolumn{4}{l}{\textbf{Valid}} & \multicolumn{4}{l}{\cellcolor[HTML]{EFEFEF}\textbf{Test}} & \multicolumn{4}{l}{\textbf{Forget}} \\ \cmidrule(l){2-13} 
\multirow{-2}{*}{\textbf{Model}} & \textbf{Recall} & \textbf{Prec} & \textbf{NDCG} & \textbf{MAP} & \cellcolor[HTML]{EFEFEF}\textbf{Recall} & \cellcolor[HTML]{EFEFEF}\textbf{Prec} & \cellcolor[HTML]{EFEFEF}\textbf{NDCG} & \cellcolor[HTML]{EFEFEF}\textbf{MAP} & \textbf{Recall} & \textbf{Prec} & \textbf{NDCG} & \textbf{MAP} \\ \midrule
Dataset & \multicolumn{12}{c}{Baby} \\ \midrule
\textsc{Mgcn} & 0.0928 & 0.0049 & 0.0419 & 0.0274 & \cellcolor[HTML]{EFEFEF}0.0941 & \cellcolor[HTML]{EFEFEF}0.0052 & \cellcolor[HTML]{EFEFEF}0.0411 & \cellcolor[HTML]{EFEFEF}0.0255 & 0.5923 & 0.1430 & 0.4705 & 0.3003 \\
\textsc{Gold} & 0.0929 & 0.0049 & 0.0413 & 0.0266 & \cellcolor[HTML]{EFEFEF}0.0944 & \cellcolor[HTML]{EFEFEF}0.0052 & \cellcolor[HTML]{EFEFEF}0.0410 & \cellcolor[HTML]{EFEFEF}0.0253 & 0.0105 & 0.0032 & 0.0072 & 0.0024 \\
\textsc{AmUn} & 0.0614 & 0.0033 & 0.0269 & 0.0170 & \cellcolor[HTML]{EFEFEF}0.0618 & \cellcolor[HTML]{EFEFEF}0.0034 & \cellcolor[HTML]{EFEFEF}0.0275 & \cellcolor[HTML]{EFEFEF}0.0174 & 0.0101 & 0.0023 & 0.0065 & 0.0025 \\
\name & 0.0889 & 0.0047 & 0.0406 & 0.0266 & \cellcolor[HTML]{EFEFEF}0.0870 & \cellcolor[HTML]{EFEFEF}0.0048 & \cellcolor[HTML]{EFEFEF}0.0384 & \cellcolor[HTML]{EFEFEF}0.0241 & 0.0105 & 0.0023 & 0.0087 & 0.0043 \\ \midrule
Dataset & \multicolumn{12}{c}{Sports} \\ \midrule
\textsc{Mgcn} & 0.1054 & 0.0056 & 0.0473 & 0.0306 & \cellcolor[HTML]{EFEFEF}0.1074 & \cellcolor[HTML]{EFEFEF}0.0060 & \cellcolor[HTML]{EFEFEF}0.0474 & \cellcolor[HTML]{EFEFEF}0.0295 & 0.4624 & 0.1091 & 0.3382 & 0.1924 \\
\textsc{Gold} & 0.1059 & 0.0056 & 0.0476 & 0.0309 & \cellcolor[HTML]{EFEFEF}0.1076 & \cellcolor[HTML]{EFEFEF}0.0060 & \cellcolor[HTML]{EFEFEF}0.0481 & \cellcolor[HTML]{EFEFEF}0.0305 & 0.0048 & 0.0015 & 0.0034 & 0.0012 \\
\textsc{AmUn} & 0.0507 & 0.0027 & 0.0220 & 0.0138 & \cellcolor[HTML]{EFEFEF}0.0537 & \cellcolor[HTML]{EFEFEF}0.0030 & \cellcolor[HTML]{EFEFEF}0.0235 & \cellcolor[HTML]{EFEFEF}0.0145 & 0.0028 & 0.0009 & 0.0020 & 0.0007 \\
\name & 0.1035 & 0.0055 & 0.0469 & 0.0307 & \cellcolor[HTML]{EFEFEF}0.1042 & \cellcolor[HTML]{EFEFEF}0.0058 & \cellcolor[HTML]{EFEFEF}0.0467 & \cellcolor[HTML]{EFEFEF}0.0296 & 0.0049 & 0.0011 & 0.0035 & 0.0014 \\ \midrule
Dataset & \multicolumn{12}{c}{Clothing} \\ \midrule
\textsc{Mgcn} & 0.0899 & 0.0046 & 0.0400 & 0.0260 & \cellcolor[HTML]{EFEFEF}0.0898 & \cellcolor[HTML]{EFEFEF}0.0047 & \cellcolor[HTML]{EFEFEF}0.0406 & \cellcolor[HTML]{EFEFEF}0.0266 & 0.8057 & 0.1785 & 0.6443 & 0.4721 \\
\textsc{Gold} & 0.0895 & 0.0045 & 0.0394 & 0.0254 & \cellcolor[HTML]{EFEFEF}0.0891 & \cellcolor[HTML]{EFEFEF}0.0046 & \cellcolor[HTML]{EFEFEF}0.0409 & \cellcolor[HTML]{EFEFEF}0.0271 & 0.0048 & 0.0011 & 0.0031 & 0.0012 \\
\textsc{AmUn} & 0.0405 & 0.0021 & 0.0176 & 0.0112 & \cellcolor[HTML]{EFEFEF}0.0415 & \cellcolor[HTML]{EFEFEF}0.0022 & \cellcolor[HTML]{EFEFEF}0.0180 & \cellcolor[HTML]{EFEFEF}0.0114 & 0.0052 & 0.0011 & 0.0039 & 0.0018 \\
\name & 0.0716 & 0.0036 & 0.0318 & 0.0206 & \cellcolor[HTML]{EFEFEF}0.0737 & \cellcolor[HTML]{EFEFEF}0.0038 & \cellcolor[HTML]{EFEFEF}0.0330 & \cellcolor[HTML]{EFEFEF}0.0214 & 0.0053 & 0.0011 & 0.0046 & 0.0024 \\ \bottomrule
\end{tabular}%
}
\caption{Unlearning 5\% of \underline{\textbf{users}}. \name matches \textsc{Gold} across validation, test, and forget sets on varied datasets outperforming \textsc{AmUn} by 49.85\%\% in unlearning generalizability, by 46.93\% in retention fidelity and high unlearning specificity.
%
}
\label{tab:userUnlearning1}
\end{table*}
\begin{table*}[t]
\centering
\resizebox{.88\textwidth}{!}{%
\begin{tabular}{@{}llllHHllHHllHH@{}}
\toprule
 &
  Set &
  \multicolumn{4}{l}{Valid} &
  \multicolumn{2}{l}{\cellcolor[HTML]{EFEFEF}Test} &
  \multicolumn{4}{l}{Forget} \\ \cmidrule(l){3-14} 
\multirow{-2}{*}{\begin{tabular}[c]{@{}l@{}}Model\\ Metric\end{tabular}} &
  K &
  Recall &
  Prec &
  NDCG &
  MAP &
  \cellcolor[HTML]{EFEFEF}Recall &
  \cellcolor[HTML]{EFEFEF}Prec &
  NDCG &
  MAP &
  Recall &
  Prec &
  NDCG &
  MAP \\ \midrule
\multicolumn{2}{l}{Dataset} &
  \multicolumn{12}{c}{Baby} \\ \midrule
\textsc{Mgcn} &
   &
  $0.9481 \pm 0.0005$ &
  $0.0452 \pm 0.0004$ &
  $0.2373 \pm 0.0006$ &
  $0.0606 \pm 0.0003$ &
  \cellcolor[HTML]{EFEFEF}$0.9073 \pm 0.0003$ &
  \cellcolor[HTML]{EFEFEF}$0.0465 \pm 0.0005$ &
  $0.1257 \pm 0.0004$ &
  $0.0090 \pm 0.0003$ &
  $0.8070 \pm 0.0005$ &
  $0.0082 \pm 0.0003$ &
  $0.0302 \pm 0.0004$ &
  $0.0007 \pm 0.0006$ \\
\textsc{Gold} &
   &
  $0.9481 \pm 0.0006$ &
  $0.0446 \pm 0.0003$ &
  $0.1924 \pm 0.0007$ &
  $0.0326 \pm 0.0004$ &
  \cellcolor[HTML]{EFEFEF}$0.9073 \pm 0.0004$ &
  \cellcolor[HTML]{EFEFEF}$0.0460 \pm 0.0006$ &
  $0.1233 \pm 0.0005$ &
  $0.0085 \pm 0.0004$ &
  $0.5103 \pm 0.0003$ &
  $0.0014 \pm 0.0001$ &
  $0.0038 \pm 0.0001$ &
  $0.0001 \pm 0.0000$ \\
\textsc{AmUn} &
   &
  $0.9271 \pm 0.0004$ &
  $0.0054 \pm 0.0006$ &
  $0.1288 \pm 0.0003$ &
  $0.0068 \pm 0.0005$ &
  \cellcolor[HTML]{EFEFEF}$0.8943 \pm 0.0007$ &
  \cellcolor[HTML]{EFEFEF}$0.0055 \pm 0.0006$ &
  $0.1081 \pm 0.0005$ &
  $0.0037 \pm 0.0004$ &
  $0.5639 \pm 0.0005$ &
  $0.0019 \pm 0.0000$ &
  $0.0000 \pm 0.0000$ &
  $0.0000 \pm 0.0000$ \\
\name &
  \multirow{-5}{*}{500} &
  $0.9481 \pm 0.0004$ &
  $0.0104 \pm 0.0005$ &
  $0.3110 \pm 0.0003$ &
  $0.1232 \pm 0.0006$ &
  \cellcolor[HTML]{EFEFEF}$0.9073 \pm 0.0003$ &
  \cellcolor[HTML]{EFEFEF}$0.0113 \pm 0.0005$ &
  $0.1172 \pm 0.0004$ &
  $0.0075 \pm 0.0006$ &
  $0.5086 \pm 0.0005$ &
  $0.0010 \pm 0.0001$ &
  $0.0000 \pm 0.0000$ &
  $0.0000 \pm 0.0000$ \\ \midrule
\multicolumn{2}{l}{Dataset} &
  \multicolumn{12}{c}{Sports} \\ \midrule
\textsc{Mgcn} &
   &
  $0.9180 \pm 0.0009$ &
  $0.0108 \pm 0.0010$ &
  $0.1768 \pm 0.0006$ &
  $0.0251 \pm 0.0001$ &
  \cellcolor[HTML]{EFEFEF}$0.8621 \pm 0.0005$ &
  \cellcolor[HTML]{EFEFEF}$0.0099 \pm 0.0006$ &
  $0.1859 \pm 0.0007$ &
  $0.0293 \pm 0.0009$ &
  $0.7573 \pm 0.0008$ &
  $0.0016 \pm 0.0001$ &
  $0.0000 \pm 0.0000$ &
  $0.0000 \pm 0.0000$ \\
\textsc{Gold} &
   &
  $0.9180 \pm 0.0006$ &
  $0.0099 \pm 0.0008$ &
  $0.1736 \pm 0.0009$ &
  $0.0237 \pm 0.0001$ &
  \cellcolor[HTML]{EFEFEF}$0.8621 \pm 0.0007$ &
  \cellcolor[HTML]{EFEFEF}$0.0096 \pm 0.0008$ &
  $0.2118 \pm 0.0009$ &
  $0.0435 \pm 0.0008$ &
  $0.6469 \pm 0.0006$ &
  $0.0013 \pm 0.0001$ &
  $0.0000 \pm 0.0000$ &
  $0.0000 \pm 0.0000$ \\
\textsc{AmUn} &
   &
  $0.9433 \pm 0.0005$ &
  $0.0045 \pm 0.0007$ &
  $0.1960 \pm 0.0006$ &
  $0.0316 \pm 0.0001$ &
  \cellcolor[HTML]{EFEFEF}$0.9013 \pm 0.0008$ &
  \cellcolor[HTML]{EFEFEF}$0.0046 \pm 0.0006$ &
  $0.2477 \pm 0.0009$ &
  $0.0660 \pm 0.0008$ &
  $0.6471 \pm 0.0007$ &
  $0.0017 \pm 0.0008$ &
  $0.0000 \pm 0.0000$ &
  $0.0000 \pm 0.0000$ \\
\name &
  \multirow{-5}{*}{1000} &
  $0.9373 \pm 0.0007$ &
  $0.0046 \pm 0.0008$ &
  $0.2327 \pm 0.0006$ &
  $0.0549 \pm 0.0001$ &
  \cellcolor[HTML]{EFEFEF}$0.9001 \pm 0.0005$ &
  \cellcolor[HTML]{EFEFEF}$0.0047 \pm 0.0001$ &
  $0.2044 \pm 0.0007$ &
  $0.0364 \pm 0.0006$ &
  $0.6693 \pm 0.0008$ &
  $0.0024 \pm 0.0006$ &
  $0.0000 \pm 0.0000$ &
  $0.0000 \pm 0.0000$ \\ \midrule
\multicolumn{2}{l}{Dataset} &
  \multicolumn{12}{c}{Clothing} \\ \midrule
\textsc{Mgcn} &
   &
  $0.9857 \pm 0.0010$ &
  $0.0079 \pm 0.0012$ &
  $0.0000 \pm 0.0000$ &
  $0.0000 \pm 0.0000$ &
  \cellcolor[HTML]{EFEFEF}$0.9690 \pm 0.0009$ &
  \cellcolor[HTML]{EFEFEF}$0.0060 \pm 0.0001$ &
  $0.0086 \pm 0.0000$ &
  $0.0002 \pm 0.0000$ &
  $0.8840 \pm 0.0009$ &
  $0.0063 \pm 0.0001$ &
  $0.0000 \pm 0.0000$ &
  $0.0000 \pm 0.0000$ \\
\textsc{Gold} &
   &
  $0.9857 \pm 0.0011$ &
  $0.0416 \pm 0.0001$ &
  $0.0000 \pm 0.0000$ &
  $0.0000 \pm 0.0000$ &
  \cellcolor[HTML]{EFEFEF}$0.9690 \pm 0.0011$ &
  \cellcolor[HTML]{EFEFEF}$0.0423 \pm 0.0006$ &
  $0.0000 \pm 0.0000$ &
  $0.0000 \pm 0.0000$ &
  $0.7134 \pm 0.0009$ &
  $0.0020 \pm 0.0001$ &
  $0.0000 \pm 0.0000$ &
  $0.0000 \pm 0.0000$ \\
\textsc{AmUn} &
   &
  $0.9721 \pm 0.0012$ &
  $0.0034 \pm 0.0001$ &
  $0.0000 \pm 0.0000$ &
  $0.0000 \pm 0.0000$ &
  \cellcolor[HTML]{EFEFEF}$0.9516 \pm 0.0008$ &
  \cellcolor[HTML]{EFEFEF}$0.0030 \pm 0.0001$ &
  $0.0796 \pm 0.0005$ &
  $0.0016 \pm 0.0000$ &
  $0.7605 \pm 0.0009$ &
  $0.0018 \pm 0.0001$ &
  $0.0000 \pm 0.0000$ &
  $0.0000 \pm 0.0000$ \\
\name &
  \multirow{-5}{*}{1500} &
  $0.9761 \pm 0.0008$ &
  $0.0034 \pm 0.0001$ &
  $0.0000 \pm 0.0000$ &
  $0.0000 \pm 0.0000$ &
  \cellcolor[HTML]{EFEFEF}$0.9526 \pm 0.0008$ &
  \cellcolor[HTML]{EFEFEF}$0.0029 \pm 0.0001$ &
  $0.0630 \pm 0.0007$ &
  $0.0009 \pm 0.0000$ &
  $0.7002 \pm 0.0008$ &
  $0.0016 \pm 0.0000$ &
  $0.0000 \pm 0.0000$ &
  $0.0000 \pm 0.0000$ \\ \bottomrule
\end{tabular}%
}
\caption{Unlearning 5\% of \underline{\textbf{items}}. \name matches \textsc{Gold} across validation, test, and forget sets on varied datasets outperforming \textsc{AmUn} by 10.83\% in unlearning specificity, and comparable retention fidelity, and unlearning generalizability.
}
\label{tab:itemUnlearning1}
\end{table*}
\begin{table*}[t]
\centering
\resizebox{.88\textwidth}{!}{%
\begin{tabular}{@{}llllllllllllll@{}}
\toprule
Model & Set & \multicolumn{4}{l}{Valid} & \multicolumn{4}{l}{\cellcolor[HTML]{EFEFEF}Test} & \multicolumn{4}{l}{Forget} \\ \cmidrule(l){2-14} 
 & @K & Recall & Precision & NDCG & MAP & \cellcolor[HTML]{EFEFEF}Recall & \cellcolor[HTML]{EFEFEF}Precision & \cellcolor[HTML]{EFEFEF}NDCG & \cellcolor[HTML]{EFEFEF}MAP & Recall & Precision & NDCG & MAP \\ \midrule
\multicolumn{2}{l}{Dataset} & \multicolumn{12}{c}{Baby} \\ \midrule
 & 20 & 0.1529 & 0.0032 & 0.0533 & 0.0286 & \cellcolor[HTML]{EFEFEF}0.1574 & \cellcolor[HTML]{EFEFEF}0.0035 & \cellcolor[HTML]{EFEFEF}0.0538 & \cellcolor[HTML]{EFEFEF}0.0274 & 0.0048 & 0.0005 & 0.0030 & 0.0011 \\
\multirow{-2}{*}{\textsc{Mgcn}} & 3200 & 0.8447 & 0.0030 & 0.2263 & 0.0497 & \cellcolor[HTML]{EFEFEF}0.8180 & \cellcolor[HTML]{EFEFEF}0.0030 & \cellcolor[HTML]{EFEFEF}0.1452 & \cellcolor[HTML]{EFEFEF}0.0096 & 0.1792 & 0.0050 & 0.3857 & 0.1195 \\
 & 20 & 0.1560 & 0.0033 & 0.0530 & 0.0277 & \cellcolor[HTML]{EFEFEF}0.1585 & \cellcolor[HTML]{EFEFEF}0.0035 & \cellcolor[HTML]{EFEFEF}0.0543 & \cellcolor[HTML]{EFEFEF}0.0279 & 0.0014 & 0.0002 & 0.0010 & 0.0004 \\
\multirow{-2}{*}{\textsc{Gold}} & 3200 & 0.8398 & 0.0030 & 0.2342 & 0.0553 & \cellcolor[HTML]{EFEFEF}0.8211 & \cellcolor[HTML]{EFEFEF}0.0029 & \cellcolor[HTML]{EFEFEF}0.1534 & \cellcolor[HTML]{EFEFEF}0.0120 & 0.2071 & 0.0049 & 0.3385 & 0.0843 \\
 & 20 & 0.1606 & 0.0034 & 0.0546 & 0.0284 & \cellcolor[HTML]{EFEFEF}0.1628 & \cellcolor[HTML]{EFEFEF}0.0036 & \cellcolor[HTML]{EFEFEF}0.0557 & \cellcolor[HTML]{EFEFEF}0.0285 & 0.0014 & 0.0002 & 0.0007 & 0.0001 \\
\multirow{-2}{*}{\textsc{AmUn}} & 3200 & 0.8417 & 0.0029 & 0.2020 & 0.0342 & \cellcolor[HTML]{EFEFEF}0.8119 & \cellcolor[HTML]{EFEFEF}0.0030 & \cellcolor[HTML]{EFEFEF}0.1490 & \cellcolor[HTML]{EFEFEF}0.0107 & 0.3416 & 0.0051 & 0.3676 & 0.1092 \\
 & 20 & 0.1479 & 0.0031 & 0.0526 & 0.0288 & \cellcolor[HTML]{EFEFEF}0.1521 & \cellcolor[HTML]{EFEFEF}0.0034 & \cellcolor[HTML]{EFEFEF}0.0525 & \cellcolor[HTML]{EFEFEF}0.0271 & 0.0014 & 0.0002 & 0.0008 & 0.0002 \\
\multirow{-2}{*}{\name} & 3200 & 0.8520 & 0.0029 & 0.3327 & 0.1426 & \cellcolor[HTML]{EFEFEF}0.8270 & \cellcolor[HTML]{EFEFEF}0.0028 & \cellcolor[HTML]{EFEFEF}0.1574 & \cellcolor[HTML]{EFEFEF}0.0134 & 0.2048 & 0.0051 & 0.7874 & 0.5434 \\ \midrule
\multicolumn{2}{l}{Dataset} & \multicolumn{12}{c}{Sports} \\ \midrule
 & 20 & 0.1681 & 0.0035 & 0.0604 & 0.0332 & \cellcolor[HTML]{EFEFEF}0.1739 & \cellcolor[HTML]{EFEFEF}0.0039 & \cellcolor[HTML]{EFEFEF}0.0612 & \cellcolor[HTML]{EFEFEF}0.0321 & 0.0030 & 0.0003 & 0.0018 & 0.0006 \\
\multirow{-2}{*}{\textsc{Mgcn}} & 7200 & 0.7392 & 0.0017 & 0.1882 & 0.0262 & \cellcolor[HTML]{EFEFEF}0.7154 & \cellcolor[HTML]{EFEFEF}0.0017 & \cellcolor[HTML]{EFEFEF}0.1941 & \cellcolor[HTML]{EFEFEF}0.0293 & 0.1034 & 0.0026 & 0.2308 & 0.0525 \\
 & 20 & 0.1694 & 0.0036 & 0.0607 & 0.0334 & \cellcolor[HTML]{EFEFEF}0.1745 & \cellcolor[HTML]{EFEFEF}0.0039 & \cellcolor[HTML]{EFEFEF}0.0617 & \cellcolor[HTML]{EFEFEF}0.0326 & 0.0012 & 0.0001 & 0.0007 & 0.0002 \\
\multirow{-2}{*}{\textsc{Gold}} & 7200 & 0.7362 & 0.0015 & 0.1844 & 0.0243 & \cellcolor[HTML]{EFEFEF}0.7133 & \cellcolor[HTML]{EFEFEF}0.0017 & \cellcolor[HTML]{EFEFEF}0.2236 & \cellcolor[HTML]{EFEFEF}0.0475 & 0.1130 & 0.0026 & 0.2271 & 0.0499 \\
 & 20 & 0.1716 & 0.0036 & 0.0608 & 0.0330 & \cellcolor[HTML]{EFEFEF}0.1752 & \cellcolor[HTML]{EFEFEF}0.0039 & \cellcolor[HTML]{EFEFEF}0.0612 & \cellcolor[HTML]{EFEFEF}0.0318 & 0.0012 & 0.0001 & 0.0006 & 0.0002 \\
\multirow{-2}{*}{\textsc{AmUn}} & 7200 & 0.7291 & 0.0017 & 0.1821 & 0.0231 & \cellcolor[HTML]{EFEFEF}0.7191 & \cellcolor[HTML]{EFEFEF}0.0017 & \cellcolor[HTML]{EFEFEF}0.1957 & \cellcolor[HTML]{EFEFEF}0.0302 & 0.1843 & 0.0026 & 0.2236 & 0.0475 \\
 & 20 & 0.1667 & 0.0035 & 0.0612 & 0.0346 & \cellcolor[HTML]{EFEFEF}0.1720 & \cellcolor[HTML]{EFEFEF}0.0038 & \cellcolor[HTML]{EFEFEF}0.0619 & \cellcolor[HTML]{EFEFEF}0.0334 & 0.0012 & 0.0001 & 0.0006 & 0.0002 \\
\multirow{-2}{*}{\name} & 7200 & 0.7449 & 0.0015 & 0.2051 & 0.0356 & \cellcolor[HTML]{EFEFEF}0.7268 & \cellcolor[HTML]{EFEFEF}0.0016 & \cellcolor[HTML]{EFEFEF}0.2120 & \cellcolor[HTML]{EFEFEF}0.0399 & 0.1376 & 0.0024 & 0.2030 & 0.0343 \\ \midrule
\multicolumn{2}{l}{Dataset} & \multicolumn{12}{c}{Clothing} \\ \midrule
 & 20 & 0.1371 & 0.0028 & 0.0491 & 0.0272 & \cellcolor[HTML]{EFEFEF}0.1360 & \cellcolor[HTML]{EFEFEF}0.0028 & \cellcolor[HTML]{EFEFEF}0.0496 & \cellcolor[HTML]{EFEFEF}0.0278 & 0.0059 & 0.0005 & 0.0032 & 0.0010 \\
\multirow{-2}{*}{\textsc{Mgcn}} & 7800 & 0.8290 & 0.0012 & 0.0604 & 0.0003 & \cellcolor[HTML]{EFEFEF}0.8213 & \cellcolor[HTML]{EFEFEF}0.0011 & \cellcolor[HTML]{EFEFEF}0.1029 & \cellcolor[HTML]{EFEFEF}0.0018 & 0.0332 & 0.0017 & 0.4999 & 0.3332 \\
 & 20 & 0.1390 & 0.0028 & 0.0497 & 0.0275 & \cellcolor[HTML]{EFEFEF}0.1355 & \cellcolor[HTML]{EFEFEF}0.0028 & \cellcolor[HTML]{EFEFEF}0.0492 & \cellcolor[HTML]{EFEFEF}0.0274 & 0.0014 & 0.0001 & 0.0007 & 0.0002 \\
\multirow{-2}{*}{\textsc{Gold}} & 7800 & 0.8197 & 0.0010 & 0.0388 & 0.0001 & \cellcolor[HTML]{EFEFEF}0.8144 & \cellcolor[HTML]{EFEFEF}0.0012 & \cellcolor[HTML]{EFEFEF}0.0696 & \cellcolor[HTML]{EFEFEF}0.0004 & 0.0502 & 0.0016 & 0.4999 & 0.3332 \\
 & 20 & 0.1372 & 0.0028 & 0.0491 & 0.0272 & \cellcolor[HTML]{EFEFEF}0.1365 & \cellcolor[HTML]{EFEFEF}0.0028 & \cellcolor[HTML]{EFEFEF}0.0497 & \cellcolor[HTML]{EFEFEF}0.0278 & 0.0013 & 0.0001 & 0.0005 & 0.0001 \\
\multirow{-2}{*}{\textsc{AmUn}} & 7800 & 0.8229 & 0.0012 & 0.0629 & 0.0003 & \cellcolor[HTML]{EFEFEF}0.8189 & \cellcolor[HTML]{EFEFEF}0.0012 & \cellcolor[HTML]{EFEFEF}0.0995 & \cellcolor[HTML]{EFEFEF}0.0015 & 0.1541 & 0.0019 & 0.4305 & 0.2499 \\
 & 20 & 0.1374 & 0.0028 & 0.0496 & 0.0278 & \cellcolor[HTML]{EFEFEF}0.1366 & \cellcolor[HTML]{EFEFEF}0.0029 & \cellcolor[HTML]{EFEFEF}0.0501 & \cellcolor[HTML]{EFEFEF}0.0282 & 0.0014 & 0.0001 & 0.0008 & 0.0002 \\
\multirow{-2}{*}{\name} & 7800 & 0.8293 & 0.0012 & 0.0502 & 0.0002 & \cellcolor[HTML]{EFEFEF}0.8126 & \cellcolor[HTML]{EFEFEF}0.0012 & \cellcolor[HTML]{EFEFEF}0.1054 & \cellcolor[HTML]{EFEFEF}0.0020 & 0.0757 & 0.0018 & 0.4305 & 0.2499 \\ \bottomrule
\end{tabular}%
}
\caption{Unlearning 5\% of \underline{\textbf{users and items}}. \name matches \textsc{Gold} across validation, test, and forget sets outperforming \textsc{AmUn} by 41.32\% in unlearning specificity and comparable unlearning generalizability and retention fidelity.
}
\label{tab:userItemUnlearning1}
\end{table*}

\section{Proposed \name Method}
\label{sec:method}
Traditionally, there are two stages for $M(\cdot, \varphi)$. In training, $M$ encodes collaborative signals inherent in the user-item interaction matrix $\mathcal{Y}$. The optimization process minimizes the BPR loss \cite{rendle2012bpr}:
\begin{align}
\label{eq:train}
    \min\mathcal{L}_\text{BPR}^{\D} &= \sum_{(u,i) \in \D} \sum_{\substack{i \in \mathcal{Y}_u^+ \\ j \in \mathcal{Y}_u^-}} -\ln \sigma(\hat{y}_{u,i} - \hat{y}_{u,j})
\end{align}
When the RS receives an unlearning request, it must first nullify the interaction data in the interaction matrix by setting $y_{u,i} = 0$ for all $y_{u,i} \in \mathcal{Y}_u^+$. Then, it would typically retrain the model $M(\cdot, \varphi)$ using the updated interaction matrix $\mathcal{Y}$. However, retraining is computationally expensive and impractical for frequent unlearning requests. Therefore, we propose unlearning the trained model $M(\cdot, \varphi)$ using \name. The overall process is illustrated in Fig.~\ref{fig:reco_unlearning}.\par







\textbf{Reverse BPR Objective.} To achieve the objectives of unlearning in MMRS, \name employs a reverse objective inspired by the concept of amnesiac unlearning~\cite{graves2021amnesiac}. This approach suggests selectively undoing the learning steps associated with a forget set $\D_f$, essentially removing the parameter updates related to forgotten data points and thus minimizing the predicted probability to a sufficiently small value. This works well for traditional machine learning tasks like image classification. Extending it to our context of MMRS, the reverse objective of user $u$ for her marked interactions is: minimize the predicted score of marked interactions relative to items with which she did not interact.
\begin{equation}
    \min\mathcal{L}_\text{RPR}^{\D_f} = \sum_{(u,i) \in \D_f} \sum_{\substack{i \in \mathcal{Y}_u^+ \\ j \in \mathcal{Y}_u^-}} \ln \sigma(\hat{y}_{u,i} - \hat{y}_{u,j})  \label{eq:RPR}
\end{equation}
However, updating the model with this reverse objective risks catastrophic forgetting of data in the retain set $\D_r$, possibly leading to inaccurate recommendations. Thus, it's essential to balance retention fidelity with unlearning specificity. \name addresses this by using the original BPR objective to preserve the retain set $\D_r$: $\mathcal{L}_\text{BPR}^{\D_r}$.


\textbf{Unlearning Multi-Modal Data} requires minimizing the predicted probability based on the discriminability of features. In addition to the collaborative signals from the user-item interaction matrix, $\mathcal{Y}$, $M(\cdot, \varphi)$ encodes item-item semantic correlations, into embeddings $\textbf{E}_{u}$ for users and $\textbf{E}_{i,m}$ for items in each modality $m \in \mathcal{M}$. A self-supervised auxiliary task maximizes mutual information between behavior features and fused multi-modal features, promoting the exploration of both, alongside $L_2$ regularization.
\begin{align}
\mathcal{L}_C^{\D} = & \sum_{u \in \D} -\log \left( \frac{\exp\left(\mathbf{e}_{u,\text{mul}} \cdot \mathbf{\bar{e}}_{u,\text{id}} / \tau \right)}{\sum_{v \in \mathcal{U}} \exp\left(\mathbf{e}_{v,\text{mul}} \cdot \mathbf{\bar{e}}_{v,\text{id}} / \tau \right)} \right) \nonumber \\
& + \sum_{i \in \D} -\log \left( \frac{\exp\left(\mathbf{e}_{i,\text{mul}} \cdot \mathbf{\bar{e}}_{i,\text{id}} / \tau \right)}{\sum_{j \in \mathcal{I}} \exp\left(\mathbf{e}_{j,\text{mul}} \cdot \mathbf{\bar{e}}_{j,\text{id}} / \tau \right)} \right)
\end{align}
, where $\tau$ is temperature. For unlearning, we introduce a negated contrastive auxiliary loss and a regularization term to \textit{reduce} the impact of learned item-item semantic correlations:
\begin{align}
\label{eq:impair}
    \mathcal{L}_\text{r}^{\D_f} &= \mathcal{L}_\text{RPR}^{\D_f} - (\lambda_C \mathcal{L}_C^{\D_f} + \lambda_\varphi \|\varphi\|^2)
\end{align}
To mitigate the risk of catastrophic forgetting, \name employs the original training objective to \textit{preserve} the knowledge within the retain set $\D_r$:
\begin{align}
\label{eq:repair}
    \mathcal{L}_\text{p}^{\D_r} &= \mathcal{L}_\text{BPR}^{\D_r} + \lambda_C \mathcal{L}_C^{\D_r} + \lambda_\varphi \|\varphi\|^2
\end{align}
However, this preservation introduces additional epochs of contrastive auxiliary loss and $L_2$ regularization, which may cause overfitting on the retain set $\D_r$. Thus, the loss in eq.\ref{eq:impair} also serves to counterbalance these effects by discouraging the model from maintaining invariance in these features.
Finally, the loss function becomes
\begin{align}
\label{eq:loss}
    \mathcal{L} = \alpha \cdot \mathcal{L}_\mathrm{p}^{\D_r} + (1 - \alpha) \cdot \mathcal{L}_\mathrm{r}^{\D_f}
\end{align}
where $\alpha$ is a hyper-parameter that determines the relative importance of preservation and reduction.

\begin{prop}
\textbf{Bayesian Interpretation of \name}. The \name objective function in Eq.~\ref{eq:RPR} can be interpreted through the Bayes theorem. Let the learning process be regarded as maximizing the posterior distribution estimated by $\varphi$, i.e., $\max P(\varphi | \D)$, with a certain prior distribution of $g(\varphi)$. Maximizing the posterior distribution $\log P(\varphi | \D_r)$ is equivalent to the RPR objective in \name, which is to minimize the RPR objective with a regularizer. 
At the optimal point, $\mathcal{L}_\text{BPR} \approx 0$ and $\left\| \frac{\partial \mathcal{L}_\text{BPR}}{\partial \varphi_0} \right\|^2 \approx 0$. Then the approximation of optimal posterior distribution is
\begin{equation}
    \mathcal{L} \approx \frac{1}{2} (\varphi - \varphi_0)^T \frac{\partial^2 \mathcal{L}_\text{BPR}}{\partial \varphi_0^2} (\varphi - \varphi_0) \label{eq:5}
\end{equation}
\end{prop}

The Kullback-Leibler (KL) divergence~\cite{kullback1951information,golatkar2020eternal,tarun2023deep} is difficult to implement in RS because the typical user-item interaction data are not inherently probabilistic but based on scores or ratings. Further, KL Divergence can be undefined when the target distribution has zero probability for an event that the input distribution considers likely, which can be problematic in sparse and high-dimensional recommendation datasets. Therefore, we \textit{propose} BPR divergence, $\beta$ as an alternative which directly measures the difference between the predicted scores of user-item interactions from different model states using mean squared error, which aligns well with the numerical nature of recommendation scores.
\begin{align}
    \beta(\varphi, \varphi') &= \frac{1}{|\mathcal{Y}_u^+|} \sum_{i \in \mathcal{Y}_u^+} \Delta y_{u,i}^2 + \frac{1}{|\mathcal{Y}_u^-|} \sum_{j \in \mathcal{Y}_u^-} \Delta y_{u,j}^2
\end{align}
where, $\Delta y_{u,i} = \hat{y}_{u,i} - \hat{y}_{u,i}^{'}
    \quad \text{and} \quad
    \Delta y_{u,j} = \hat{y}_{u,j} - \hat{y}_{u,j}^{'}$.


\begin{prop}
\textbf{Information Bound of \name}. Let the posterior distribution before and after unlearning the forget set $\D_f$, be $P(\varphi | \D_T)$ and $P(\varphi | \D_r)$, [since $\D_T - \D_f = \D_r$], respectively. 
The convergence between can be expressed as $\beta(\varphi, \varphi') \leq \epsilon$ where $n$ denote the number of epochs during unlearning. Since $\epsilon \propto 1/n$, we have $\epsilon = k \times \frac{1}{n}$, where $k$ is a constant.
\end{prop}

As shown in Figure~\ref{fig:bprdiv}, the BPR divergence between the gold and unlearned models on the forget set $\D_f$ decreases with more epochs, indicating that less information about $\D_f$ remains in the model over time.

\textbf{Time Complexity } for unlearning LightGCN with BPR and auxiliary contrastive loss using \name is $T = \mathcal{O} \left( \mathcal{M} \cdot d \cdot |\E| \cdot d_m + |\mathcal{U}| \cdot |\mathcal{I}| \cdot K + d^2 + |\mathcal{M}| \cdot d \right)$, where $\mathcal{M}$ is the number of modalities, $d$ is the embedding dimension, $|\E|$ is the number of edges, $d_m$ is the feature dimension for modality $m$, $|\mathcal{U}|$ is the number of users, $|\mathcal{I}|$ is the number of items, and $K$ is the number of negative samples per user-item pair. 

\section{Experiments and Results}
\label{sec:exp}
\subsection{Experimental Setup}
\textbf{Datasets and baselines.} We select \textsc{Mgcn}~\cite{yu2023multi} as our base model for several reasons: \ding{182} It is the state-of-the-art MMRS with BPR. \ding{183} \textsc{Mgcn} represents BPR-based models well, making improvements likely to generalize. \ding{184} Unlike traditional methods like Collaborative Filtering and Matrix Factorization, it handles multi-modal features through graph convolutional representations. \ding{185} Other loss functions for uni-modal systems don't address multi-modal complexities, and those that do, perform weaker than \textsc{Mgcn}. So, focusing on \textsc{Mgcn} ensures our approach is relevant.

We use three distinct categories within the Amazon dataset~\cite{hou2024bridging} that are used to  benchmark MMRS.
The model trained on the retain set (named as the \textsc{Gold} model) from scratch is used as baseline. We prepare another baseline by re-purposing the image classification unlearning~\textsc{AmUn}~\cite{graves2021amnesiac} for MMRS. 

Previous RS unlearning methods using matrix factorization, collaborative filtering, or sequential systems operate on behavior matrices, making them incompatible with graph-based architectures. The code for approximate methods \textsc{Rrl}, \textsc{Ifru}, and \textsc{Scif} is unavailable. 
Exact methods, \textsc{RecEraser} and \textsc{UltraRE}, require partitioning, which is challenging due to the need for specialized handling of each modality and ensuring coherence during aggregation, adding complexity outside the scope of this work.

\textbf{Evaluation metrics}
To evaluate a given model $M(\cdot, \varphi)$, we need \ding{182} model predictions: ranked list of user-item pairs; \ding{183} the ground truth: user-item interactions $\mathcal{Y}$; and, \ding{184} $K$: number of the top recommendations to consider. 
To capture a variety of aspects of performance, we use \ding{182} predictive metrics, that reflect the ``correctness", how well $M(\cdot, \varphi)$ finds relevant items such as Recall at K and Precision at K; and \ding{183} ranking metrics, that reflect ranking quality, how well $M(\cdot, \varphi)$ can sort items from more relevant to less relevant. NDCG considers both relevance and position of items in ranked list. MAP measures the average Precision across different Recall levels for a ranked list.



\textbf{Settings and Hyper-parameters}
All experiments are performed on $4$x NVIDIA RTX2080 ($32$GB). We use the identical settings as in the case of \textsc{Mgcn} for training the original model $M(\cdot, \varphi)$. The data interaction history of each user is split $8:1:1$ for training, validation and testing. The training set is used to create retain set and forget set. We initialize the embedding with Xavier initialization of dimension $64$, set the regularization coefficient $\lambda_\varphi = 10^{-4}$, and batch size $B = 2048$. For the self-supervised task, we set the temperature to $0.2$. For convergence, the early stopping and total epochs are fixed at $20$ and $1000$, respectively. We use Recall@20 on the validation data as the training-stopping indicator following \cite{zhang2022latent}. We conducted experiments five times using different seed values and report the standard deviation values for item unlearning in Table~\ref{tab:itemUnlearning1}. For the other types of unlearning, the standard deviation values were found to be negligible and are therefore omitted.

%
\begin{table*}[h]
\centering
\resizebox{0.9\textwidth}{!}{%
\begin{tabular}{@{}lcccccccccccccc@{}}
\toprule
$\mathbf{\alpha}$ & \textbf{0.001} & \textbf{0.003} & \textbf{0.01} & \textbf{0.03} & \textbf{0.1} & \textbf{0.2} & \textbf{0.3} & \textbf{0.4} & \textbf{0.5} & \textbf{0.6} & \textbf{0.7} & \textbf{0.8} & \textbf{0.9} \\ \midrule
\textbf{Valid} & 0.0779 & 0.0851 & 0.0882 & 0.0887 & 0.0893 & 0.0893 & 0.0913 & 0.0909 & 0.0897 & 0.0880 & 0.0878 & 0.0804 & 0.0775 \\
\textbf{Test} & 0.0748 & 0.0827 & 0.0848 & 0.0856 & 0.0871 & 0.0892 & 0.0920 & 0.0915 & 0.0910 & 0.0885 & 0.0885 & 0.0839 & 0.0809 \\
\textbf{Forget} & 0.0104 & 0.0103 & 0.0107 & 0.0102 & 0.0105 & 0.0103 & 0.0105 & 0.0106 & 0.0105 & 0.0107 & 0.0104 & 0.0107 & 0.0102 \\ \bottomrule
\end{tabular}
}
\caption{Tuning $\alpha$ for balance. Recall@20 on validation, test, and forget sets while unlearning 5\% of users in the Baby dataset. Lower $\alpha$ values optimize recall and convergence, while higher values improve forgetting at the expense of retain set recall.}
\label{tab:alpha}
\end{table*}

\begin{table}[h!]
\centering
\resizebox{\columnwidth}{!}{
\begin{tabular}{lll}
\toprule
\textbf{Metric} & \textbf{User-Centric} & \textbf{Item-Centric} \\
\midrule
$\text{Recall}@K$ & $\frac{|\text{Relevant}_u \cap \text{Retrieved}_u|}{|\text{Relevant}_u|}$ & $\frac{\sum_{j=1}^{K} \text{rel}_{j,i}}{|\text{Users}_i|}$ \\
$\text{Recall}@K$ & $\frac{1}{|\mathcal{U}|} \sum_{u \in \mathcal{U}} \text{Recall}_{u}@K$ & $\frac{1}{|\mathcal{I}|} \sum_{i \in \mathcal{I}} \text{Recall}_{i}@K$ \\
$\text{Prec}@K$ & $\frac{|Rel_u \cap Rec_u|}{|Rec_u|}$ & $\text{Prec}_{i}@K = \frac{\sum_{j=1}^{K} \text{rel}_{j,i}}{K}$ \\
$\text{Precision}@K$ & $\frac{1}{|\mathcal{U}|} \sum_{u \in \mathcal{U}} \text{Prec}_{u}@K$ & $\frac{1}{|\mathcal{I}|} \sum_{i \in \mathcal{I}} \text{Prec}_{i}@K$ \\
$\text{DCG}@K$ & $\sum_{i=1}^{K} \frac{2^{rel_{i,u}} - 1}{\log_{2}(i+1)}$ & $\sum_{j=1}^{K} \frac{2^{rel_{j,i}} - 1}{\log_{2}(j+1)}$ \\
$\text{IDCG}@K$ & $\sum_{i=1}^{K} \frac{1}{\log_{2}(i+1)}$ & $\sum_{j=1}^{K} \frac{1}{\log_{2}(j+1)}$ \\
$\text{NDCG}@K$ & $\frac{\text{DCG}_{u}@K}{\text{IDCG}_{u}@K}$ & $\frac{\text{DCG}_{i}@K}{\text{IDCG}_{i}@K}$ \\
$\text{NDCG}@K$ & $\frac{1}{|\mathcal{U}|} \sum_{u \in \mathcal{U}} \text{NDCG}_{u}@K$ & $\frac{1}{|\mathcal{I}|} \sum_{i \in \mathcal{I}} \text{NDCG}_{i}@K$ \\
$\text{AP}@N$ & $\sum_{k=1}^{N}\frac{P(k) \cdot \text{rel}_{u}(k)}{\min(m, N)}$ & $\sum_{k=1}^{N}\frac{P(k) \cdot \text{rel}_{i}(k)}{\min(m, N)}$ \\
$\text{MAP}@N$ & $\frac{1}{|\mathcal{U}|} \sum_{u \in \mathcal{U}} \text{AP}_{u}@N$ & $\frac{1}{|\mathcal{I}|} \sum_{i \in \mathcal{I}} \text{AP}_{i}@N$ \\
\bottomrule
\end{tabular}
}
\caption{User-Centric vs. Item-Centric Metrics. Measures relevance from the user's perspective versus from the item's perspective, independent of users recommended to.}
\label{tab:metrics}
\end{table}

\begin{figure}[t]
    \centering
    \begin{subfigure}[b]{0.48\columnwidth}
        \includegraphics[width=\textwidth]{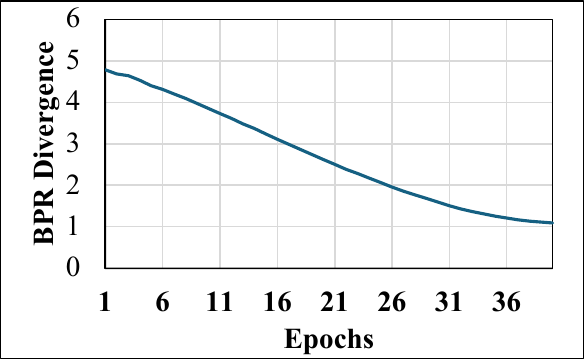}
        \caption{BPR Divergence Over Epochs. Decreases with more epochs, indicating reduced retention of $\D_f$ information.}
        \label{fig:bprdiv}
    \end{subfigure}
    \begin{subfigure}[b]{0.50\columnwidth}
        \includegraphics[width=\textwidth]{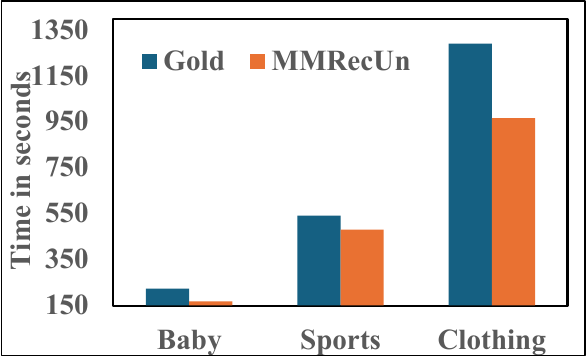}
        \caption{\name reduces time by up to 1.3× compared to retraining from scratch, saving 25\% of the time.}
        \label{fig:timesaved}
    \end{subfigure}
    \caption{Information bound and efficiency analysis.}
    \label{fig:figure2}
\end{figure}

\subsection{Results}
\label{sec:Results}
We conduct experiments to assess the unlearning performance of \name in the four aspects. First, we compare recall, precision, NDCG and MAP metrics on \textit{forget set}. Values closer to the \textsc{Gold} model signify better \textit{Unlearning Specificity}. Second, we compare the metrics on \textit{test set}. Values closer to the \textsc{Gold} model signify better \textit{Retention Fidelity}. Third, we compare the metrics on \textit{validation set}. Values closer to the \textsc{Gold} model signify better \textit{Unlearning Generalizability}. Finally, we compare the time duration required for unlearning to compare \textit{efficiency}.

\underline{\textit{User-centric metrics}} measure the relevance from user's perspective i.e., whether the relevant items (the user likes or interacts with) are present in the top $K$ recommendations. Users typically expect to see a small number of recommendations (e.g., $5$ or $10$), hence we take $K = 5, 10, 20, 50$ to assess how well the system can provide personalized and focused recommendations. Larger values of $K$, such as $20$ or $50$, are useful for evaluating the system's ability to cover a wider range of user preferences or measure performance in suggesting diverse options and catering to different user tastes. These metrics primarily focus on user satisfaction and do not capture the effectiveness of forgetting specific items. Hence, we \textit{introduce} several \textit{item-centric metrics}.

\underline{\textit{Item-centric metrics}} (Table~\ref{tab:metrics}) measure the relevance of the recommended items themselves i.e., how well the recommended items meet unlearning criteria, regardless of which users receive them. We take $K = 500, 1000, 1500$ to ensure that all items in the forget set $\D_f$ are considered for evaluating performance.
\textbf{User Unlearning.} Table~\ref{tab:userUnlearning1} shows the results of unlearning $5\%$ users. The \textsc{Gold} model, which has never seen the forget set, falls in performance on this set as expected. However, it retains good performance on the test and validation sets. On the validation set, \name consistently achieves better recall scores, outperforming \textsc{AmUn} by $29.6\%$, $49.85\%$, and $34.75\%$ on the three datasets (baby, sports, and clothing, respectively), showcasing \textit{best} unlearning generalizability. On the test set, \name maintains closest recall scores, surpassing \textsc{AmUn} by $26.69\%$, $46.93\%$, and $36.13\%$, demonstrating \textit{best} retention fidelity. On the forget set, \name achieves better recall scores, outperforming \textsc{AmUn} by $3.8\%$ and $43.75\%$ on two datasets and slightly falling back by $2.08\%$ on the clothing dataset, indicating \textit{high} unlearning specificity. Similar trends follow for other metrics. 

\textbf{Item Unlearning} Table~\ref{tab:itemUnlearning1} shows the results of unlearning $5\%$ items. On the validation set, the recall scores differ by $2.21\%$, $0.65\%$, and $0.41\%$ on the three datasets, showcasing \textit{comparable} unlearning generalizability. Similarly, on the test set, the recall scores differ by $1.43\%$, $0.13\%$, and $0.10\%$, demonstrating \textit{comparable} retention fidelity. However, \name achieves good recall scores, outperforming \textsc{AmUn} by $10.83\%$ and $8.45\%$ on two datasets and slightly falling back by $3.43\%$ on the sports dataset, indicating \textit{high} unlearning specificity. 

\textbf{User and Item Unlearning.} Table~\ref{tab:userItemUnlearning1} presents the results of unlearning both $5\%$ users and $5\%$ items. On \textit{user-centric metrics}, the performance of \textsc{AmUn} and \name is comparable. The recall scores differ by $2.89\%$, $1.83\%$ and $0\%$ approximately across three datasets on the validation set, test set and forget set, respectively, showcasing \textit{comparable} unlearning generalizability, retention fidelity and unlearning specificity. On \textit{item-centric metrics}, \name achieves recall score up to $41.32\%$ higher than \textsc{AmUn} on forget set, indicating \textit{high} unlearning specificity. The recall scores differ by $2.14\%$ and $1.83\%$ approximately across three datasets on the validation and test sets, respectively, showcasing \textit{comparable} unlearning generalizability and retention fidelity. We repeat the experiment 5 times using different seed values and report the standard deviation in Table~\ref{tab:itemUnlearning1}. For the other types of unlearning, the standard deviation were found to be negligible and are therefore omitted.

\textbf{Unlearning efficiency.} On the baby dataset, \name takes $169.41$ seconds in contrast to $225.73$ seconds of \textsc{Gold} model. Similarly, on the clothing dataset, \name takes $968.96$ seconds in contrast to $1294.44$ seconds of \textsc{Gold} model. Thus, with \name, we experience up to $1.3\times$ accelerated unlearning than retraining from scratch (Figure~\ref{fig:timesaved}). In other words, $25\%$ time is saved.

\textbf{Tuning hyper-parameter $\mathbf{\alpha}$.} We assess how $\alpha$ balances adaptation and reform in unlearning, testing values from $0.001$ to $0.9$ (Table~\ref{tab:alpha}). Lower $\alpha$ values ($0.1$-$0.2$) effectively unlearn the forget set with minimal impact on retain set recall, and require fewer epochs for convergence, indicating an optimal balance. As $\alpha$ increases ($0.3$-$0.9$), the model emphasizes more on the reform loss, needing more epochs to forget, adversely affecting retain set recall. Extremely low $\alpha$ values ($<0.01$) hinder effective forgetting, also increasing epochs and negatively impacting retain set. The optimal $\alpha$ depends on the specific task's forgetting severity and time constraints: higher $\alpha$ may ensure stronger forgetting with minor recall loss, while lower $\alpha$ enables faster convergence.

\section{Conclusion}
\name is the first framework for MMRS unlearning that handles complex multi-modal data and item unlearning. It caters to diverse unlearning requests, like user preference adjustments, item removal etc.; and scenarios: user, item and user-item simultaneous unlearning. We define three key properties to measure MMRS unlearning; introduce item-centric metrics with traditional user-centric ones; and propose BPR divergence as a robust alternative to KL divergence to compare recommendation scores. Our experiments demonstrate \name's superiority in retention fidelity, unlearning specificity, generalizability, and efficiency.

{ \section*{Acknowledgments}
This research/project is supported by the National Research Foundation, Singapore under its Strategic Capability Research Centres Funding Initiative. Any opinions, findings and conclusions or recommendations expressed in this material are those of the author(s) and do not reflect the views of National Research Foundation, Singapore.}
\medskip
{
 \small
 \bibliography{aaai25} 
}
\newpage
\appendix
\twocolumn[
]
\section{Appendix}
\subsection{Related Work}
\label{sec:related}
\textbf{Recommender Systems.} Recommender systems can be categorized into various types based on their underlying techniques. These include collaborative filtering (CF), matrix factorization (MF), content-based filtering (CBF), sequential recommender systems (SRS), graph convolutional network-based (GCN) and hybrids of these systems. Recent advancements have seen a surge in research on multi-modal recommender systems ~\cite{xu2018graphcar,wei2019mmgcn,wei2020graph,liu2023multimodal}, which leverage multiple modalities such as text, images, and videos to enhance recommendation accuracy and user satisfaction. The current state-of-the-art \textsc{Mgcn}~\cite{yu2023multi}, a GCN model, leverages multiple views to separate modality features and behavior features, enhancing feature discriminability, and considers the relative importance of different modalities for improved user preference modeling.\par
\textbf{Machine Unlearning.} In recent years, significant advancements have been made in the field of machine unlearning~\cite{cao2015towards,bourtoule2021machine,sinha2024unstar,chundawat2024conda, chatterjee2024unified, sharma2024unlearning} across various domains such as image classification~\cite{tarun2023fast,chundawat2023can,chundawat2023zero}, regression~\cite{tarun2023deep}, federated learning~\cite{wu2022federated}, graph learning~\cite{chen2022graph,wang2023inductive}, and more. Exact unlearning~\cite{bourtoule2021machine} aims to completely remove the influence of specific data points from the model through algorithmic-level retraining. The advantage is the model will behave as if the unlearned data had never been seen. While providing strong guarantees of removal, exact unlearning usually demands extensive computational resources and is primarily suitable for simpler models.~\textit{Approximate unlearning} focuses on efficiently minimizing the influence of target data points through limited parameter-level updates to the model. While not removing influence entirely, approximate unlearning significantly reduces computational and time costs. It enables practical unlearning applications even for large-scale and complex machine learning models, where exact retraining is infeasible.\par
\textbf{Multi-modal Machine Unlearning.} In \textsc{Mmul}~\cite{cheng2023multimodal}, unlearning is accomplished by minimizing the dissimilarity between the multi-modal representations of deleted pairs and randomly drawn unassociated pairs. At the same time, uni-modal and multi-modal knowledge is preserved by minimizing the difference between the uni-modal and multi-modal representations produced by the unlearned model and the original model.

\textbf{Recommendation Unlearning.} Various methods for unlearning have been explored in different types of recommender systems as summarized in Table~\ref{tab:related}. 
For collaborative filtering, \textsc{Caboose}~\cite{schelter2023forget} incrementally adjusts the user-item matrices and re-calculates similarities only for affected entries to reflect the removal of a specific interaction. This approach avoids complete re-computation, effectively conserving computational resources. \textsc{Laser}~\cite{li2024making} partitions users according to collaborative embedding and trains them sequentially with curriculum learning. 

For matrix factorization, \textsc{IMCorrect}~\cite{liu2023recommendation} adjusts the interaction matrix by weakening item pairs' similarity if they had high similarity due to similar interaction patterns in the removed interactions, reflecting the unlearned recommendations. In another work\textsc{Cmumf}~\cite{zhang2023closed}, performs a closed-form unlearning update based on the total Hessian-based Newton step, allowing for the efficient removal of the influence of specific rows or columns from the trained MF factors. \textsc{ALS}~\cite{xu2022netflix} modifies the intermediate confidence matrix used in Alternating Least Squares
to achieve fast forgetting. \textsc{AltEraser}~\cite{liu2022forgetting} unlearns via alternating optimization in Alternating Least squares.\par




For sequential recommender systems, \textsc{DyRand}~\cite{ye2023sequence} leverages label noise injection to make random predictions for sequences to be unlearned. Among uni-modal GCN-based systems, exact unlearning has been investigated by \textsc{RecEraser}~\cite{chen2022recommendation}. It partitions training set into shards, train sub-models and employs attention-based adaptive aggregation strategies to combine performance of shards. For unlearning, one of the sub-models whose shard contains the interaction to be unlearned and the aggregation part needs to be retrained. \textsc{UltraRE}~\cite{li2023ultrare} improves partitioning by optimal balanced clustering and aggregation by logistic regression. 


Among uni-modal GCN-based systems, approximate unlearning has been explored by \textsc{Ifru}~\cite{zhang2023recommendation} which updates the model based on influence of data to be unlearned. \textsc{Scif}~\cite{li2023selective} improves by selectively updating user embeddings based on their collaborative influence. \textsc{Rrl}~\cite{you2024rrl} uses reverse BPR loss to unlearn and Fisher Information Matrix to preserve remaining user-item interactions. Though, computing the inverse of Hessian for influence function and Fisher Information Matrix are computationally expensive operations.
Other related work includes federated unlearning for on-device recommendation~\cite{yuan2023federated}, attribute unlearning~\cite{li2023making,chen2024post},~\cite{ganhor2022unlearning}, unlearning session-based recommendations~\cite{xin2024effectiveness}, and unlearning LLMs fine-tuned for recommendation~\cite{wang2024towards}.

\subsection{Challenges in Multi-modal Recommendation Unlearning}
\label{sec:challenges}
Previous methods cannot be directly applied to multi-modal GCN-based recommender systems due to the following challenges. 

\textbf{Divergence in Model Representations}: Multi-modal GCN models, incorporate heterogeneous user-item image, text and behavior information into a unified graph, contrasting with the matrices, latent factors, feature representations, and temporal sequences used in traditional recommender\begin{landscape}
\begin{table}[tb]
\centering
\caption{Comparison of Different Recommendation Unlearning Methods.}
\label{tab:related}
\begin{tabular}{@{}p{2.2cm}p{1.6cm}p{2.8cm}p{4.1cm}p{1.5cm}p{5cm}p{1.5cm}p{0.5cm}@{}}
\toprule
\multicolumn{1}{l}{\textbf{Method}} &
  \multicolumn{1}{l}{\textbf{\shortstack[l]{Recomm-\\ender Type}}} &
  \multicolumn{1}{l}{\textbf{Models}} &
  \multicolumn{1}{l}{\textbf{Modality}} &
  \multicolumn{1}{l}{\textbf{\shortstack[l]{Unlearning\\ Request}}} &
  \multicolumn{1}{l}{\textbf{Procedure}} &
  \multicolumn{1}{l}{\textbf{\shortstack[l]{Evaluation\\ Metrics}}} &
  \multicolumn{1}{l}{\textbf{\shortstack[l]{Unlearning\\Paradigm}}} \\ \midrule
\textsc{Laser}~\cite{li2024making} &
  CF &
  DMF, NMF &
  U-I Behavior Matrix &
  Interactions &
  Partition users acc. to collaborative embedding \& train sequentially with curriculum learning &
  Hit Ratio, NDCG &
  Exact \\
\textsc{Caboose}~\cite{schelter2023forget} &
  CF &
  kNN &
  U-I Behavior Matrix &
  Interactions &
  Re-calculate affected entries in   user-item matrix &
  Recall &
  Approx. \\
\textsc{IMCorrect}~\cite{liu2023recommendation} &
  CF, MF &
  SLIM, GF-CF, MF, AutoRec &
  U-I, U-U, U-I Behavior Matrix &
  Interactions &
  Weakening item pairs' similarity   in the interaction matrix &
  Recall &
  Approx. \\
\textsc{ALS}~\cite{xu2022netflix} &
  MF &
  MF &
  U-I Behavior Matrix &
  Interactions &
  modifies the intermediate confidence matrix used in Alternating Least Squares to unlearn &
  AUC &
  Exact \\
\textsc{AltEraser}~\cite{liu2022forgetting} &
  MF &
  NMF &
  U-I Behavior Matrix &
  Interactions &
  via alternating optimization in Alternating Least squares &
  Recall, NDCG, RBO &
  Approx. \\
\textsc{Cmumf}~\cite{zhang2023closed} &
  MF &
  ALS, PF, CGD, Mom-SGD, RMSProp, BALM &
  U-I Behavior Matrix &
  Users, Items &
  update based on the total Hessian-based Newton step &
  Normalized norm difference &
  Exact \\
\textsc{DyRand}~\cite{ye2023sequence} &
  SRS &
  SASRec, S3-Rec, STOSA &
  sequential, temporal, session-based U-I behavior Matrix &
  Interaction sequence &
  label noise injection to make random predictions &
  Hit Ratio, NDCG, MRR &
  Approx. \\
\textsc{RecEraser}~\cite{chen2022recommendation} &
  MF, GCN &
  BPR, WMF, LightGCN &
  U-I Behavior Matrix \& Bi-partite Graph &
  Interactions &
  partition training set into shards and train sub-models &
  Recall, NDCG &
  Exact \\
\textsc{UltraRE}~\cite{li2023ultrare} &
  MF, GCN &
  DMF, LightGCN &
  U-I Behavior Matrix \& Bi-partite Graph &
  Interactions &
  improves partitioning by optimal balanced clustering and aggregation by logistic regression &
  Hit Ratio, NDCG &
  Exact \\
\textsc{Rrl}~\cite{you2024rrl} &
  MF, GCN &
  BPR, LightGCN &
  U-I Behavior Matrix \& Bi-partite Graph &
  Interactions &
  reverse BPR loss to unlearn \& Fisher Information Matrix to preserve &
  Recall, NDCG &
  Approx. \\
\textsc{Ifru}~\cite{zhang2023recommendation} &
  MF, GCN &
  MF, LightGCN &
  U-I Behavior Matrix \& Bi-partite Graph &
  Interactions &
  update based on influence of data to be unlearned &
  AUC &
  Approx. \\
\textsc{Scif}~\cite{li2023selective} &
  MF, GCN &
  NMF, LightGCN &
  U-I Behavior Matrix \& Bi-partite Graph &
  Interactions &
  selectively updating user embedding based on collaborative influence &
  Hit Ratio, NDCG &
  Approx. \\ \midrule
\name \textit{(Ours)} &
  MM-GCN &
  MGCN &
  U-I Behavior Bi-partite Graph, user \& item features: text, image, audio, video, location, biometric etc. &
  Interactions, Users, Items &
  Impair and repair with BPR Loss &
  Precision, Recall &
  Approx. \\ \bottomrule
\end{tabular}%
\end{table}
\end{landscape} systems like CF, MF, CBF, and SRS, respectively. Consequently, transferring unlearning mechanisms is not as straight-forward.\par
    
\textbf{Incompatibility with Uni-modal Approaches}: Uni-modal recommender systems, even those based on GCNs, fail to directly apply to multi-modal GCN-based systems due to the inherent complexities of integrating diverse modalities~\cite{cheng2023multimodal}. Both the graph structure and feature embeddings are tightly integrated, making it non-trivial to modify one component without affecting the other. Unlearning interactions in one modality may have cascading effects on the representations and relationships within the entire graph.\par

\par\textbf{Expensive operations}: The computational overhead associated with computing the Hessian matrix or adjusting the interaction matrix or computing the inverse of Hessian for influence function or Fisher Information Matrix may become prohibitive as the size of the dataset or the number of interactions increases, making some of the methods less practical for large-scale recommender systems.\par

\textbf{Ill-effects of sharding}: \ding{182} Partitioning data breaks the graph structure. Partitioning based on items restricts the system's ability to recommend items across different partitions to users, and partitioning based on users leads to insufficient training data and poor data heterogeneity~\cite{koch2023no}. This causes a significant drop ($\sim 30\%$) in performance~\cite{you2024rrl}. \ding{183} This adds unlearning overhead of extra computational cost and time at the time of training . \ding{184} When items that need to be forgotten come from multiple shards, the worst case efficiency falls to the level of retraining from scratch. \ding{185} The model's initial setup, partitioned based on users during training, poses a challenge to implementing item-related data unlearning, as it cannot be readily reconfigured to partition data based on items. \ding{186} Item-related data unlearning concurrently with user-related data unlearning is not possible.\par

\par\textbf{Burden of aggregation}: \ding{182} Additional steps to aggregate performance across shards introduce significant overhead costs, with the burden increasing proportionally to the number of shards~\cite{ramezani2021learn}. \ding{183} This adds unlearning overhead of extra computational cost and time at the time of inference. \ding{184} The partition-aggregation framework is prohibitive for the models which have already been trained.\par

These constraints underscore the need for tailored methodologies specifically designed for unlearning in multi-modal recommender systems.

\begin{table*}[h]
\centering
\caption{Unlearning $\mathbf{5\%}$ \underline{\textbf{users}} in forget set across three datasets: Baby, Sports and Clothing. Recall, Precision, NDCG and MAP scores with $K=5, 10, 20, 50$ on validation, test and forget set. 
}
\label{tab:userUnlearning2}
\resizebox{\textwidth}{!}{%
\begin{tabular}{@{}lrllllllllllll@{}}
\toprule
 &
  Set &
  \multicolumn{4}{l}{Valid} &
  \multicolumn{4}{l}{\cellcolor[HTML]{EFEFEF}Test} &
  \multicolumn{4}{l}{Forget} \\ \cmidrule(l){2-14} 
\multirow{-2}{*}{Model} &
  @K &
  Recall &
  Prec &
  NDCG &
  MAP &
  \cellcolor[HTML]{EFEFEF}Recall &
  \cellcolor[HTML]{EFEFEF}Prec &
  \cellcolor[HTML]{EFEFEF}NDCG &
  \cellcolor[HTML]{EFEFEF}MAP &
  Recall &
  Prec &
  NDCG &
  MAP \\ \midrule
\multicolumn{2}{l}{Dataset} &
  \multicolumn{12}{c}{Baby} \\ \midrule
 &
  5 &
  0.0400 &
  0.0084 &
  0.0270 &
  0.0224 &
  \cellcolor[HTML]{EFEFEF}0.0385 &
  \cellcolor[HTML]{EFEFEF}0.0084 &
  \cellcolor[HTML]{EFEFEF}0.0252 &
  \cellcolor[HTML]{EFEFEF}0.0202 &
  0.3215 &
  0.2837 &
  0.3713 &
  0.2586 \\
 &
  10 &
  0.0620 &
  0.0065 &
  0.0341 &
  0.0253 &
  \cellcolor[HTML]{EFEFEF}0.0609 &
  \cellcolor[HTML]{EFEFEF}0.0067 &
  \cellcolor[HTML]{EFEFEF}0.0325 &
  \cellcolor[HTML]{EFEFEF}0.0232 &
  0.4531 &
  0.2079 &
  0.4164 &
  0.2768 \\
 &
  20 &
  0.0928 &
  0.0049 &
  0.0419 &
  0.0274 &
  \cellcolor[HTML]{EFEFEF}0.0941 &
  \cellcolor[HTML]{EFEFEF}0.0052 &
  \cellcolor[HTML]{EFEFEF}0.0411 &
  \cellcolor[HTML]{EFEFEF}0.0255 &
  0.5923 &
  0.1430 &
  0.4705 &
  0.3003 \\
\multirow{-4}{*}{\textsc{Mgcn}} &
  50 &
  0.1555 &
  0.0033 &
  0.0545 &
  0.0294 &
  \cellcolor[HTML]{EFEFEF}0.1585 &
  \cellcolor[HTML]{EFEFEF}0.0035 &
  \cellcolor[HTML]{EFEFEF}0.0541 &
  \cellcolor[HTML]{EFEFEF}0.0275 &
  0.7666 &
  0.0795 &
  0.5315 &
  0.3210 \\ \cmidrule(l){2-14} 
 &
  5 &
  0.0384 &
  0.0081 &
  0.0260 &
  0.0216 &
  \cellcolor[HTML]{EFEFEF}0.0378 &
  \cellcolor[HTML]{EFEFEF}0.0083 &
  \cellcolor[HTML]{EFEFEF}0.0249 &
  \cellcolor[HTML]{EFEFEF}0.0200 &
  0.0034 &
  0.0041 &
  0.0046 &
  0.0023 \\
 &
  10 &
  0.0590 &
  0.0062 &
  0.0327 &
  0.0243 &
  \cellcolor[HTML]{EFEFEF}0.0596 &
  \cellcolor[HTML]{EFEFEF}0.0066 &
  \cellcolor[HTML]{EFEFEF}0.0321 &
  \cellcolor[HTML]{EFEFEF}0.0229 &
  0.0048 &
  0.0032 &
  0.0048 &
  0.0020 \\
 &
  20 &
  0.0929 &
  0.0049 &
  0.0413 &
  0.0266 &
  \cellcolor[HTML]{EFEFEF}0.0944 &
  \cellcolor[HTML]{EFEFEF}0.0052 &
  \cellcolor[HTML]{EFEFEF}0.0410 &
  \cellcolor[HTML]{EFEFEF}0.0253 &
  0.0105 &
  0.0032 &
  0.0072 &
  0.0024 \\
\multirow{-4}{*}{\textsc{Gold}} &
  50 &
  0.1577 &
  0.0033 &
  0.0542 &
  0.0286 &
  \cellcolor[HTML]{EFEFEF}0.1596 &
  \cellcolor[HTML]{EFEFEF}0.0036 &
  \cellcolor[HTML]{EFEFEF}0.0542 &
  \cellcolor[HTML]{EFEFEF}0.0274 &
  0.0193 &
  0.0024 &
  0.0102 &
  0.0027 \\ \cmidrule(l){2-14} 
 &
  5 &
  0.0248 &
  0.0052 &
  0.0165 &
  0.0135 &
  \cellcolor[HTML]{EFEFEF}0.0252 &
  \cellcolor[HTML]{EFEFEF}0.0056 &
  \cellcolor[HTML]{EFEFEF}0.0170 &
  \cellcolor[HTML]{EFEFEF}0.0139 &
  0.0033 &
  0.0031 &
  0.0040 &
  0.0021 \\
 &
  10 &
  0.0402 &
  0.0042 &
  0.0214 &
  0.0155 &
  \cellcolor[HTML]{EFEFEF}0.0405 &
  \cellcolor[HTML]{EFEFEF}0.0045 &
  \cellcolor[HTML]{EFEFEF}0.0221 &
  \cellcolor[HTML]{EFEFEF}0.0160 &
  0.0056 &
  0.0026 &
  0.0048 &
  0.0022 \\
 &
  20 &
  0.0614 &
  0.0033 &
  0.0269 &
  0.0170 &
  \cellcolor[HTML]{EFEFEF}0.0618 &
  \cellcolor[HTML]{EFEFEF}0.0034 &
  \cellcolor[HTML]{EFEFEF}0.0275 &
  \cellcolor[HTML]{EFEFEF}0.0174 &
  0.0101 &
  0.0023 &
  0.0065 &
  0.0025 \\
\multirow{-4}{*}{\textsc{AmUn}} &
  50 &
  0.1060 &
  0.0023 &
  0.0358 &
  0.0184 &
  \cellcolor[HTML]{EFEFEF}0.1068 &
  \cellcolor[HTML]{EFEFEF}0.0024 &
  \cellcolor[HTML]{EFEFEF}0.0367 &
  \cellcolor[HTML]{EFEFEF}0.0188 &
  0.0192 &
  0.0018 &
  0.0094 &
  0.0028 \\ \cmidrule(l){2-14}
 &
  5 &
  0.0395 &
  0.0083 &
  0.0265 &
  0.0219 &
  \cellcolor[HTML]{EFEFEF}0.0362 &
  \cellcolor[HTML]{EFEFEF}0.0079 &
  \cellcolor[HTML]{EFEFEF}0.0239 &
  \cellcolor[HTML]{EFEFEF}0.0193 &
  0.0056 &
  0.0045 &
  0.0069 &
  0.0041 \\
 &
  10 &
  0.0612 &
  0.0064 &
  0.0335 &
  0.0247 &
  \cellcolor[HTML]{EFEFEF}0.0571 &
  \cellcolor[HTML]{EFEFEF}0.0063 &
  \cellcolor[HTML]{EFEFEF}0.0307 &
  \cellcolor[HTML]{EFEFEF}0.0220 &
  0.0075 &
  0.0032 &
  0.0076 &
  0.0042 \\
 &
  20 &
  0.0889 &
  0.0047 &
  0.0406 &
  0.0266 &
  \cellcolor[HTML]{EFEFEF}0.0870 &
  \cellcolor[HTML]{EFEFEF}0.0048 &
  \cellcolor[HTML]{EFEFEF}0.0384 &
  \cellcolor[HTML]{EFEFEF}0.0241 &
  0.0105 &
  0.0023 &
  0.0087 &
  0.0043 \\
\multirow{-4}{*}{\name} &
  50 &
  0.1427 &
  0.0030 &
  0.0513 &
  0.0283 &
  \cellcolor[HTML]{EFEFEF}0.1458 &
  \cellcolor[HTML]{EFEFEF}0.0032 &
  \cellcolor[HTML]{EFEFEF}0.0504 &
  \cellcolor[HTML]{EFEFEF}0.0260 &
  0.0158 &
  0.0015 &
  0.0104 &
  0.0045 \\ \midrule
\multicolumn{2}{l}{Dataset} &
  \multicolumn{12}{c}{Sports} \\ \midrule
 &
  5 &
  0.0460 &
  0.0096 &
  0.0304 &
  0.0250 &
  \cellcolor[HTML]{EFEFEF}0.0453 &
  \cellcolor[HTML]{EFEFEF}0.0099 &
  \cellcolor[HTML]{EFEFEF}0.0296 &
  \cellcolor[HTML]{EFEFEF}0.0237 &
  0.2188 &
  0.1955 &
  0.2491 &
  0.1614 \\
 &
  10 &
  0.0711 &
  0.0075 &
  0.0386 &
  0.0283 &
  \cellcolor[HTML]{EFEFEF}0.0709 &
  \cellcolor[HTML]{EFEFEF}0.0078 &
  \cellcolor[HTML]{EFEFEF}0.0379 &
  \cellcolor[HTML]{EFEFEF}0.0270 &
  0.3211 &
  0.1474 &
  0.2844 &
  0.1727 \\
 &
  20 &
  0.1054 &
  0.0056 &
  0.0473 &
  0.0306 &
  \cellcolor[HTML]{EFEFEF}0.1074 &
  \cellcolor[HTML]{EFEFEF}0.0060 &
  \cellcolor[HTML]{EFEFEF}0.0474 &
  \cellcolor[HTML]{EFEFEF}0.0295 &
  0.4624 &
  0.1091 &
  0.3382 &
  0.1924 \\
\multirow{-4}{*}{\textsc{Mgcn}} &
  50 &
  0.1667 &
  0.0035 &
  0.0596 &
  0.0326 &
  \cellcolor[HTML]{EFEFEF}0.1729 &
  \cellcolor[HTML]{EFEFEF}0.0039 &
  \cellcolor[HTML]{EFEFEF}0.0606 &
  \cellcolor[HTML]{EFEFEF}0.0316 &
  0.6611 &
  0.0666 &
  0.4057 &
  0.2123 \\ \cmidrule(l){2-14} 
 &
  5 &
  0.0460 &
  0.0096 &
  0.0306 &
  0.0252 &
  \cellcolor[HTML]{EFEFEF}0.0449 &
  \cellcolor[HTML]{EFEFEF}0.0099 &
  \cellcolor[HTML]{EFEFEF}0.0301 &
  \cellcolor[HTML]{EFEFEF}0.0245 &
  0.0019 &
  0.0022 &
  0.0025 &
  0.0013 \\
 &
  10 &
  0.0710 &
  0.0075 &
  0.0388 &
  0.0286 &
  \cellcolor[HTML]{EFEFEF}0.0709 &
  \cellcolor[HTML]{EFEFEF}0.0078 &
  \cellcolor[HTML]{EFEFEF}0.0386 &
  \cellcolor[HTML]{EFEFEF}0.0279 &
  0.0037 &
  0.0022 &
  0.0032 &
  0.0012 \\
 &
  20 &
  0.1059 &
  0.0056 &
  0.0476 &
  0.0309 &
  \cellcolor[HTML]{EFEFEF}0.1076 &
  \cellcolor[HTML]{EFEFEF}0.0060 &
  \cellcolor[HTML]{EFEFEF}0.0481 &
  \cellcolor[HTML]{EFEFEF}0.0305 &
  0.0048 &
  0.0015 &
  0.0034 &
  0.0012 \\
\multirow{-4}{*}{\textsc{Gold}} &
  50 &
  0.1665 &
  0.0035 &
  0.0597 &
  0.0328 &
  \cellcolor[HTML]{EFEFEF}0.1704 &
  \cellcolor[HTML]{EFEFEF}0.0038 &
  \cellcolor[HTML]{EFEFEF}0.0608 &
  \cellcolor[HTML]{EFEFEF}0.0325 &
  0.0082 &
  0.0011 &
  0.0046 &
  0.0013 \\ \cmidrule(l){2-14} 
 &
  5 &
  0.0196 &
  0.0042 &
  0.0132 &
  0.0109 &
  \cellcolor[HTML]{EFEFEF}0.0207 &
  \cellcolor[HTML]{EFEFEF}0.0046 &
  \cellcolor[HTML]{EFEFEF}0.0140 &
  \cellcolor[HTML]{EFEFEF}0.0114 &
  0.0012 &
  0.0012 &
  0.0014 &
  0.0007 \\
 &
  10 &
  0.0323 &
  0.0034 &
  0.0173 &
  0.0126 &
  \cellcolor[HTML]{EFEFEF}0.0338 &
  \cellcolor[HTML]{EFEFEF}0.0038 &
  \cellcolor[HTML]{EFEFEF}0.0183 &
  \cellcolor[HTML]{EFEFEF}0.0131 &
  0.0017 &
  0.0011 &
  0.0016 &
  0.0007 \\
 &
  20 &
  0.0507 &
  0.0027 &
  0.0220 &
  0.0138 &
  \cellcolor[HTML]{EFEFEF}0.0537 &
  \cellcolor[HTML]{EFEFEF}0.0030 &
  \cellcolor[HTML]{EFEFEF}0.0235 &
  \cellcolor[HTML]{EFEFEF}0.0145 &
  0.0028 &
  0.0009 &
  0.0020 &
  0.0007 \\
\multirow{-4}{*}{\textsc{AmUn}} &
  50 &
  0.0891 &
  0.0019 &
  0.0297 &
  0.0150 &
  \cellcolor[HTML]{EFEFEF}0.0930 &
  \cellcolor[HTML]{EFEFEF}0.0021 &
  \cellcolor[HTML]{EFEFEF}0.0314 &
  \cellcolor[HTML]{EFEFEF}0.0157 &
  0.0051 &
  0.0007 &
  0.0028 &
  0.0008 \\ \cmidrule(l){2-14}
 &
  5 &
  0.0470 &
  0.0098 &
  0.0309 &
  0.0253 &
  \cellcolor[HTML]{EFEFEF}0.0444 &
  \cellcolor[HTML]{EFEFEF}0.0097 &
  \cellcolor[HTML]{EFEFEF}0.0295 &
  \cellcolor[HTML]{EFEFEF}0.0239 &
  0.0023 &
  0.0021 &
  0.0026 &
  0.0013 \\
 &
  10 &
  0.0701 &
  0.0074 &
  0.0384 &
  0.0284 &
  \cellcolor[HTML]{EFEFEF}0.0701 &
  \cellcolor[HTML]{EFEFEF}0.0077 &
  \cellcolor[HTML]{EFEFEF}0.0379 &
  \cellcolor[HTML]{EFEFEF}0.0273 &
  0.0037 &
  0.0017 &
  0.0031 &
  0.0013 \\
 &
  20 &
  0.1035 &
  0.0055 &
  0.0469 &
  0.0307 &
  \cellcolor[HTML]{EFEFEF}0.1042 &
  \cellcolor[HTML]{EFEFEF}0.0058 &
  \cellcolor[HTML]{EFEFEF}0.0467 &
  \cellcolor[HTML]{EFEFEF}0.0296 &
  0.0049 &
  0.0011 &
  0.0035 &
  0.0014 \\
\multirow{-4}{*}{\name} &
  50 &
  0.1631 &
  0.0035 &
  0.0589 &
  0.0326 &
  \cellcolor[HTML]{EFEFEF}0.1685 &
  \cellcolor[HTML]{EFEFEF}0.0038 &
  \cellcolor[HTML]{EFEFEF}0.0598 &
  \cellcolor[HTML]{EFEFEF}0.0317 &
  0.0083 &
  0.0008 &
  0.0046 &
  0.0015 \\ \midrule
\multicolumn{2}{l}{Dataset} &
  \multicolumn{12}{c}{Clothing} \\ \midrule
 &
  5 &
  0.0398 &
  0.0081 &
  0.0259 &
  0.0212 &
  \cellcolor[HTML]{EFEFEF}0.0399 &
  \cellcolor[HTML]{EFEFEF}0.0083 &
  \cellcolor[HTML]{EFEFEF}0.0265 &
  \cellcolor[HTML]{EFEFEF}0.0218 &
  0.4733 &
  0.3863 &
  0.5102 &
  0.3908 \\
 &
  10 &
  0.0609 &
  0.0062 &
  0.0327 &
  0.0240 &
  \cellcolor[HTML]{EFEFEF}0.0609 &
  \cellcolor[HTML]{EFEFEF}0.0063 &
  \cellcolor[HTML]{EFEFEF}0.0333 &
  \cellcolor[HTML]{EFEFEF}0.0246 &
  0.6514 &
  0.2762 &
  0.5812 &
  0.4359 \\
 &
  20 &
  0.0899 &
  0.0046 &
  0.0400 &
  0.0260 &
  \cellcolor[HTML]{EFEFEF}0.0898 &
  \cellcolor[HTML]{EFEFEF}0.0047 &
  \cellcolor[HTML]{EFEFEF}0.0406 &
  \cellcolor[HTML]{EFEFEF}0.0266 &
  0.8057 &
  0.1785 &
  0.6443 &
  0.4721 \\
\multirow{-4}{*}{\textsc{Mgcn}} &
  50 &
  0.1385 &
  0.0028 &
  0.0496 &
  0.0275 &
  \cellcolor[HTML]{EFEFEF}0.1370 &
  \cellcolor[HTML]{EFEFEF}0.0029 &
  \cellcolor[HTML]{EFEFEF}0.0501 &
  \cellcolor[HTML]{EFEFEF}0.0281 &
  0.9392 &
  0.0878 &
  0.6923 &
  0.4930 \\ \cmidrule(l){2-14} 
 &
  5 &
  0.0387 &
  0.0079 &
  0.0252 &
  0.0206 &
  \cellcolor[HTML]{EFEFEF}0.0407 &
  \cellcolor[HTML]{EFEFEF}0.0084 &
  \cellcolor[HTML]{EFEFEF}0.0272 &
  \cellcolor[HTML]{EFEFEF}0.0225 &
  0.0017 &
  0.0012 &
  0.0017 &
  0.0009 \\
 &
  10 &
  0.0601 &
  0.0061 &
  0.0321 &
  0.0234 &
  \cellcolor[HTML]{EFEFEF}0.0607 &
  \cellcolor[HTML]{EFEFEF}0.0063 &
  \cellcolor[HTML]{EFEFEF}0.0337 &
  \cellcolor[HTML]{EFEFEF}0.0251 &
  0.0025 &
  0.0010 &
  0.0021 &
  0.0010 \\
 &
  20 &
  0.0895 &
  0.0045 &
  0.0394 &
  0.0254 &
  \cellcolor[HTML]{EFEFEF}0.0891 &
  \cellcolor[HTML]{EFEFEF}0.0046 &
  \cellcolor[HTML]{EFEFEF}0.0409 &
  \cellcolor[HTML]{EFEFEF}0.0271 &
  0.0048 &
  0.0011 &
  0.0031 &
  0.0012 \\
\multirow{-4}{*}{\textsc{Gold}} &
  50 &
  0.1383 &
  0.0028 &
  0.0492 &
  0.0270 &
  \cellcolor[HTML]{EFEFEF}0.1356 &
  \cellcolor[HTML]{EFEFEF}0.0028 &
  \cellcolor[HTML]{EFEFEF}0.0501 &
  \cellcolor[HTML]{EFEFEF}0.0286 &
  0.0085 &
  0.0008 &
  0.0043 &
  0.0013 \\ \cmidrule(l){2-14} 
 &
  5 &
  0.0168 &
  0.0034 &
  0.0109 &
  0.0090 &
  \cellcolor[HTML]{EFEFEF}0.0166 &
  \cellcolor[HTML]{EFEFEF}0.0034 &
  \cellcolor[HTML]{EFEFEF}0.0110 &
  \cellcolor[HTML]{EFEFEF}0.0091 &
  0.0025 &
  0.0020 &
  0.0029 &
  0.0017 \\
 &
  10 &
  0.0263 &
  0.0027 &
  0.0140 &
  0.0102 &
  \cellcolor[HTML]{EFEFEF}0.0267 &
  \cellcolor[HTML]{EFEFEF}0.0028 &
  \cellcolor[HTML]{EFEFEF}0.0143 &
  \cellcolor[HTML]{EFEFEF}0.0104 &
  0.0033 &
  0.0014 &
  0.0032 &
  0.0017 \\
 &
  20 &
  0.0405 &
  0.0021 &
  0.0176 &
  0.0112 &
  \cellcolor[HTML]{EFEFEF}0.0415 &
  \cellcolor[HTML]{EFEFEF}0.0022 &
  \cellcolor[HTML]{EFEFEF}0.0180 &
  \cellcolor[HTML]{EFEFEF}0.0114 &
  0.0052 &
  0.0011 &
  0.0039 &
  0.0018 \\
\multirow{-4}{*}{\textsc{AmUn}} &
  50 &
  0.0703 &
  0.0014 &
  0.0234 &
  0.0121 &
  \cellcolor[HTML]{EFEFEF}0.0722 &
  \cellcolor[HTML]{EFEFEF}0.0015 &
  \cellcolor[HTML]{EFEFEF}0.0241 &
  \cellcolor[HTML]{EFEFEF}0.0124 &
  0.0093 &
  0.0009 &
  0.0053 &
  0.0020 \\ \cmidrule(l){2-14}
 &
  5 &
  0.0313 &
  0.0063 &
  0.0204 &
  0.0168 &
  \cellcolor[HTML]{EFEFEF}0.0311 &
  \cellcolor[HTML]{EFEFEF}0.0065 &
  \cellcolor[HTML]{EFEFEF}0.0209 &
  \cellcolor[HTML]{EFEFEF}0.0173 &
  0.0030 &
  0.0023 &
  0.0037 &
  0.0022 \\
 &
  10 &
  0.0486 &
  0.0049 &
  0.0260 &
  0.0191 &
  \cellcolor[HTML]{EFEFEF}0.0492 &
  \cellcolor[HTML]{EFEFEF}0.0051 &
  \cellcolor[HTML]{EFEFEF}0.0267 &
  \cellcolor[HTML]{EFEFEF}0.0197 &
  0.0040 &
  0.0017 &
  0.0041 &
  0.0023 \\
 &
  20 &
  0.0716 &
  0.0036 &
  0.0318 &
  0.0206 &
  \cellcolor[HTML]{EFEFEF}0.0737 &
  \cellcolor[HTML]{EFEFEF}0.0038 &
  \cellcolor[HTML]{EFEFEF}0.0330 &
  \cellcolor[HTML]{EFEFEF}0.0214 &
  0.0053 &
  0.0011 &
  0.0046 &
  0.0024 \\
\multirow{-4}{*}{\name} &
  50 &
  0.1155 &
  0.0023 &
  0.0405 &
  0.0220 &
  \cellcolor[HTML]{EFEFEF}0.1165 &
  \cellcolor[HTML]{EFEFEF}0.0024 &
  \cellcolor[HTML]{EFEFEF}0.0415 &
  \cellcolor[HTML]{EFEFEF}0.0227 &
  0.0088 &
  0.0008 &
  0.0057 &
  0.0025 \\ \bottomrule
\end{tabular}%
}
\end{table*}
\begin{table*}[t]
\centering
\caption{Unlearning $\mathbf{5\%}$ \underline{\textbf{items}} in forget set across three datasets: Baby, Sports and Clothing. Recall, Precision, NDCG and MAP scores with $K=500, 1000, 1500$ on validation, test and forget set. 
}
\label{tab:itemUnlearning2}
\resizebox{\textwidth}{!}{%
\begin{tabular}{@{}llllllllllllll@{}}
\toprule
 &
  Set &
  \multicolumn{4}{l}{Valid} &
  \multicolumn{4}{l}{\cellcolor[HTML]{EFEFEF}Test} &
  \multicolumn{4}{l}{Forget} \\ \cmidrule(l){3-14} 
\multirow{-2}{*}{\begin{tabular}[c]{@{}l@{}}Model\\ Metric\end{tabular}} &
  K &
  Recall &
  Prec &
  NDCG &
  MAP &
  \cellcolor[HTML]{EFEFEF}Recall &
  \cellcolor[HTML]{EFEFEF}Prec &
  \cellcolor[HTML]{EFEFEF}NDCG &
  \cellcolor[HTML]{EFEFEF}MAP &
  Recall &
  Prec &
  NDCG &
  MAP \\ \midrule
\multicolumn{2}{l}{Dataset} &
  \multicolumn{12}{c}{Baby} \\ \midrule
\textsc{Mgcn} &
   &
  0.9481 &
  0.0452 &
  0.2373 &
  0.0606 &
  \cellcolor[HTML]{EFEFEF}0.9073 &
  \cellcolor[HTML]{EFEFEF}0.0465 &
  \cellcolor[HTML]{EFEFEF}0.1257 &
  \cellcolor[HTML]{EFEFEF}0.0090 &
  0.8070 &
  0.0082 &
  0.0302 &
  0.0007 \\
\textsc{Gold} &
   &
  0.9481 &
  0.0446 &
  0.1924 &
  0.0326 &
  \cellcolor[HTML]{EFEFEF}0.9073 &
  \cellcolor[HTML]{EFEFEF}0.0460 &
  \cellcolor[HTML]{EFEFEF}0.1233 &
  \cellcolor[HTML]{EFEFEF}0.0085 &
  0.5103 &
  0.0014 &
  0.0038 &
  0.0001 \\
\textsc{AmUn} &
   &
  0.9271 &
  0.0054 &
  0.1288 &
  0.0068 &
  \cellcolor[HTML]{EFEFEF}0.8943 &
  \cellcolor[HTML]{EFEFEF}0.0055 &
  \cellcolor[HTML]{EFEFEF}0.1081 &
  \cellcolor[HTML]{EFEFEF}0.0037 &
  0.5639 &
  0.0019 &
  0.0000 &
  0.0000 \\
\textsc{Scif} &
   &
  0.0000 &
  0.0000 &
  0.0000 &
  0.0000 &
  \cellcolor[HTML]{EFEFEF}0.0000 &
  \cellcolor[HTML]{EFEFEF}0.0000 &
  \cellcolor[HTML]{EFEFEF}0.0000 &
  \cellcolor[HTML]{EFEFEF}0.0000 &
  0.0000 &
  0.0000 &
  0.0000 &
  0.0000 \\
\name &
  \multirow{-5}{*}{500} &
  0.9481 &
  0.0104 &
  0.3110 &
  0.1232 &
  \cellcolor[HTML]{EFEFEF}0.9073 &
  \cellcolor[HTML]{EFEFEF}0.0113 &
  \cellcolor[HTML]{EFEFEF}0.1172 &
  \cellcolor[HTML]{EFEFEF}0.0075 &
  0.5086 &
  0.0010 &
  0.0000 &
  0.0000 \\ \midrule
\multicolumn{2}{l}{Dataset} &
  \multicolumn{12}{c}{Sports} \\ \midrule
\textsc{Mgcn} &
   &
  0.9180 &
  0.0108 &
  0.1768 &
  0.0251 &
  \cellcolor[HTML]{EFEFEF}0.8621 &
  \cellcolor[HTML]{EFEFEF}0.0099 &
  \cellcolor[HTML]{EFEFEF}0.1859 &
  \cellcolor[HTML]{EFEFEF}0.0293 &
  0.7573 &
  0.0016 &
  0.0000 &
  0.0000 \\
\textsc{Gold} &
   &
  0.9180 &
  0.0099 &
  0.1736 &
  0.0237 &
  \cellcolor[HTML]{EFEFEF}0.8621 &
  \cellcolor[HTML]{EFEFEF}0.0096 &
  \cellcolor[HTML]{EFEFEF}0.2118 &
  \cellcolor[HTML]{EFEFEF}0.0435 &
  0.6469 &
  0.0013 &
  0.0000 &
  0.0000 \\
\textsc{AmUn} &
   &
  0.9433 &
  0.0045 &
  0.1960 &
  0.0316 &
  \cellcolor[HTML]{EFEFEF}0.9013 &
  \cellcolor[HTML]{EFEFEF}0.0046 &
  \cellcolor[HTML]{EFEFEF}0.2477 &
  \cellcolor[HTML]{EFEFEF}0.0660 &
  0.6471 &
  0.0017 &
  0.0000 &
  0.0000 \\
\textsc{Scif} &
   &
  0.0000 &
  0.0000 &
  0.0000 &
  0.0000 &
  \cellcolor[HTML]{EFEFEF}0.0000 &
  \cellcolor[HTML]{EFEFEF}0.0000 &
  \cellcolor[HTML]{EFEFEF}0.0000 &
  \cellcolor[HTML]{EFEFEF}0.0000 &
  0.0000 &
  0.0000 &
  0.0000 &
  0.0000 \\
\name &
  \multirow{-5}{*}{1000} &
  0.9373 &
  0.0046 &
  0.2327 &
  0.0549 &
  \cellcolor[HTML]{EFEFEF}0.9001 &
  \cellcolor[HTML]{EFEFEF}0.0047 &
  \cellcolor[HTML]{EFEFEF}0.2044 &
  \cellcolor[HTML]{EFEFEF}0.0364 &
  0.6693 &
  0.0024 &
  0.0000 &
  0.0000 \\ \midrule
\multicolumn{2}{l}{Dataset} &
  \multicolumn{12}{c}{Clothing} \\ \midrule
\textsc{Mgcn} &
   &
  0.9857 &
  0.0079 &
  0.0000 &
  0.0000 &
  \cellcolor[HTML]{EFEFEF}0.9690 &
  \cellcolor[HTML]{EFEFEF}0.0060 &
  \cellcolor[HTML]{EFEFEF}0.0086 &
  \cellcolor[HTML]{EFEFEF}0.0002 &
  0.8840 &
  0.0063 &
  0.0000 &
  0.0000 \\
\textsc{Gold} &
   &
  0.9857 &
  0.0416 &
  0.0000 &
  0.0000 &
  \cellcolor[HTML]{EFEFEF}0.9690 &
  \cellcolor[HTML]{EFEFEF}0.0423 &
  \cellcolor[HTML]{EFEFEF}0.0000 &
  \cellcolor[HTML]{EFEFEF}0.0000 &
  0.7134 &
  0.0020 &
  0.0000 &
  0.0000 \\
\textsc{AmUn} &
   &
  0.9721 &
  0.0034 &
  0.0000 &
  0.0000 &
  \cellcolor[HTML]{EFEFEF}0.9516 &
  \cellcolor[HTML]{EFEFEF}0.0030 &
  \cellcolor[HTML]{EFEFEF}0.0796 &
  \cellcolor[HTML]{EFEFEF}0.0016 &
  0.7605 &
  0.0018 &
  0.0000 &
  0.0000 \\
\textsc{Scif} &
   &
  0.0000 &
  0.0000 &
  0.0000 &
  0.0000 &
  \cellcolor[HTML]{EFEFEF}0.0000 &
  \cellcolor[HTML]{EFEFEF}0.0000 &
  \cellcolor[HTML]{EFEFEF}0.0000 &
  \cellcolor[HTML]{EFEFEF}0.0000 &
  0.0000 &
  0.0000 &
  0.0000 &
  0.0000 \\
\name &
  \multirow{-5}{*}{1500} &
  0.9761 &
  0.0034 &
  0.0000 &
  0.0000 &
  \cellcolor[HTML]{EFEFEF}0.9526 &
  \cellcolor[HTML]{EFEFEF}0.0029 &
  \cellcolor[HTML]{EFEFEF}0.0630 &
  \cellcolor[HTML]{EFEFEF}0.0009 &
  0.7002 &
  0.0016 &
  0.0000 &
  0.0000 \\ \bottomrule
\end{tabular}%
}
\end{table*}
\subsection{Learning and Unlearning Workflows for Multi-modal Recommendation Systems}
\label{sec:workflows}
\textbf{Training and Inference.} In the traditional workflow, the recommender model $M(\cdot, \varphi)$ undergoes two key stages. At the time of training, the model aims to encode collaborative signals inherent in the user-item interaction matrix $\mathcal{Y}$, along with item-item semantic correlations, into embeddings $\textbf{E}_{u}$ for users and $\textbf{E}_{i,m}$ for items in each modality $m \in \mathcal{M}$.

To enhance the model's ability to capture features across different modalities accurately, a behavior-aware fuser component is integrated. This component allows for flexible fusion weight allocation based on user modality preferences derived from behavior features. Additionally, a self-supervised auxiliary task is introduced to maximize the mutual information between behavior features and fused multi-modal features, promoting the exploration of behavior and multi-modal information. Refer to \textsc{Mgcn} paper~\cite{yu2023multi} for details.

The optimization process involves minimizing the Bayesian Personalized Ranking (BPR) loss $\mathcal{L}_\text{BPR}$ \cite{rendle2012bpr}, augmented by auxiliary self-supervised tasks. This combined loss function, $\mathcal{L}$, updates the model parameters $\varphi$, with hyper-parameters $\lambda_C$ and $\lambda_\varphi$ controlling the impact of the contrastive auxiliary task and $L_2$ regularization, respectively.
\begin{align}
    \mathcal{L}_\text{BPR} &= \sum_{u \in \mathcal{U}} \sum_{\substack{i \in \mathcal{Y}_u^+ \\ j \in \mathcal{Y}_u^-}} -\ln \sigma(\hat{y}_{u,i} - \hat{y}_{u,j}) \\
\label{eq:trainSup}
    \mathcal{L} &= \mathcal{L}_\text{BPR} + \lambda_C \mathcal{L}_C + \lambda_\varphi \|\varphi\|^2
\end{align}

During inference, the model obtains embeddings $\textbf{e}_{u}$ and $\textbf{e}_{i,m}$ by passing users and items through it, $\textbf{e}_{u} = M(u, \varphi)$, $\textbf{e}_{i,m} = M(i, \varphi)$.
These enhanced features are then utilized to compute the likelihood of interaction between a user $u$ and an item $i$, represented by the inner product $\hat{y}_{u,i} = e_u^T e_i$. Finally, it recommends items to users based on these preference scores, prioritizing items with higher scores that have not been interacted with by the user.

\begin{figure}[ht]
    \centering
    \includegraphics[width=.7\columnwidth]
    {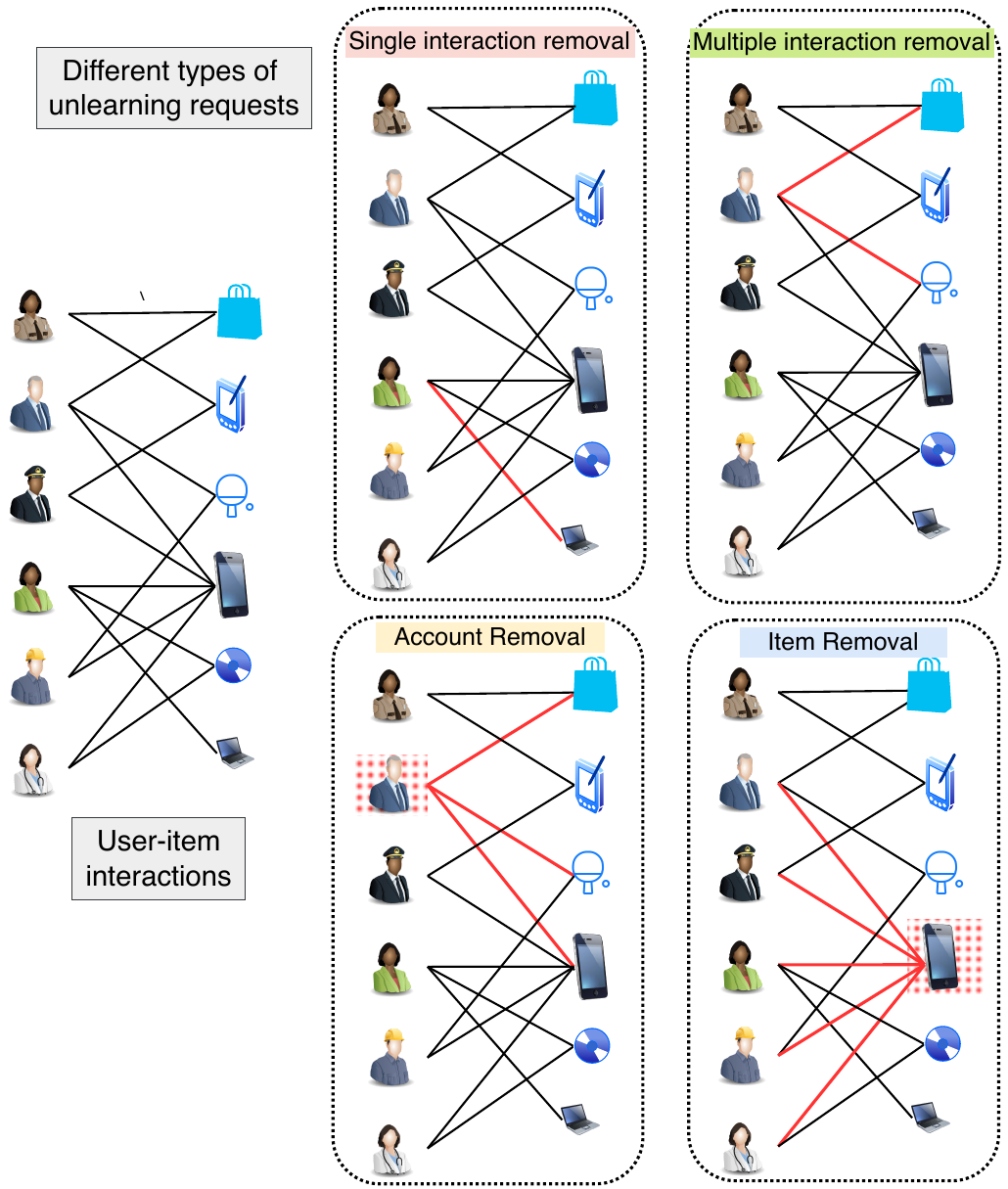}
    \caption{Unlearning request types in MMRS, addressing privacy, preferences, bias elimination, and legal compliance.}
    \label{fig:unlearning_requests}
\end{figure}

\textbf{Unlearning with Traditional Workflow: Retraining.} To cater to demands for privacy, legal compliance, and evolving user preferences and content licensing grow, the recommender model $M(\cdot, \varphi)$ has to adapt dynamically. Thus, either of the four unlearning requests can come up as shown in Fig \ref{fig:unlearning_requests}.
\begin{enumerate}
    \item Single interaction removal (privacy needs): This request involves removing a single interaction between a user and an item from the model. It addresses privacy concerns by allowing users to delete specific interactions they no longer want associated with their profile. For example, if a user wants to remove a movie they watched from their viewing history for privacy reasons, the system would unlearn this interaction.
    
    \item Multiple interaction removal (user preference adjustment, bias elimination): In this request, multiple interactions of the same user with different items or multiple interactions of the same item across many users are removed. It aims to adjust user preferences, eliminate bias, and mitigate shilling attacks~\cite{fang2018poisoning,tang2020revisiting} by updating the model based on the removal of these interactions. Shilling attacks involve artificially inflating the ratings or interactions of certain items to manipulate recommendations, and this removal process helps counteract such malicious behavior.
    
    \item Removal of all interactions of a user for account deletion (privacy laws): This request involves removing all interactions associated with a specific user from the model, effectively deleting their entire profile and history from the system to comply with privacy regulations.
    
    \item Removal of all interactions related to an item (evolving license): In this request, all interactions related to a particular item are removed from the model. It addresses evolving content licensing agreements by adapting the model to the absence or unavailability of certain items. For instance, if a music label withdraws its catalog from the platform, the system would unlearn all interactions associated with songs from that label to comply with the licensing changes.
\end{enumerate}

After receiving the requests, the recommendation system must execute two processes. The first process involves removing data from the database by nullifying interaction data in the interaction matrix, $y_{u,i} = 0$, $\forall y_{u,i} \in \mathcal{Y}_u^+$. The second process requires retraining the recommendation model $M(\cdot, \varphi)$ using the updated interaction matrix $\mathcal{Y}$, following the training process specified by Eq.~\ref{eq:train}. 

\textbf{Proposed \name Workflow for Unlearning.} As discussed in Section~\ref{sec:challenges}, retraining the recommendation model from scratch poses computational inefficiencies due to the large volume of interactions and the size of the model itself. This challenge is further exacerbated when multiple unlearning requests occur sequentially over time. Hence, we advocate for unlearning the initially trained model $M(\cdot, \varphi)$ using \name.

\begin{table}[H]
\centering
\caption{List of Symbols and Their Descriptions}
\resizebox{\columnwidth}{!}{
\begin{tabular}{@{}ll@{}}
\toprule
\textbf{Symbol} & \textbf{Description} \\ \midrule
$\mathcal{U}$ & Set of users \\
$\mathcal{I}$ & Set of items \\
$\textbf{E}_{u}$ & Embedding matrix for users \\
$\textbf{E}_{i,m}$ & Embedding matrix for each item modality \\
$d$ & Embedding dimension for users \\
$d_m$ & Dimension of the features for modality $m$ \\
$\mathcal{M}$ & Set of modalities \\
$\mathcal{Y}$ & User-item interaction matrix \\
$\G$ & Sparse behavior graph \\
$\V$ & Set of nodes in the graph \\
$\E$ & Set of edges in the graph \\
$M(\cdot, \varphi)$ & Recommendation model with parameters $\varphi$ \\
$\D$ & Dataset of user-item interactions \\
$\D_T$ & Training dataset \\
$\D_v$ & Validation dataset \\
$\D_t$ & Test dataset \\
$\D_f$ & Forget set \\
$\D_r$ & Retain set \\
$\varphi$ & Model parameters \\
$\varphi_r$ & Parameters of the gold model trained on $\D_r$ \\
$\varphi_u$ & Parameters of the unlearned model \\
$U$ & Unlearning algorithm \\ \bottomrule
\end{tabular}%
}
\label{tab:symbols}
\end{table}
\subsection{Bayesian Interpretation of \name}
\label{sec:BayesianInter}
We present the theoretical interpretation of the objective in equation~\ref{eq:RPR} through the Bayesian theorem. The learning process can be regarded as maximizing the posterior distribution estimated by $\varphi$, i.e., $\max P(\varphi | \D_T)$, with a certain prior distribution of $g(\varphi)$. Such posterior distribution $P(\varphi | \D_T)$ can be decomposed as follows.
\begin{align}
    P(\varphi | \D_r, \D_f) &= P(\varphi | \D_r)P(\D_f | \varphi, \D_r) \frac{P(\D_f, \D_r)}{P(\D_f)} \label{eq:1} \\
    \log P(\varphi | \D_T) &= \log P(\varphi | \D_r) + \log P(\D_f | \varphi, \D_r) - \log P(\D_f) \label{eq:2}
\end{align}
We can derive the log posterior distribution $\log P(\varphi | \D_r)$ as,
\begin{equation}
    \log P(\varphi | \D_r) = \log P(\varphi | \D_T) - \log P(\D_f | \varphi) + \log P(\D_f) \label{eq:3}
\end{equation}
Maximizing the log posterior distribution, \(\log P(\varphi | \D_r)\), is the same as retraining a recommendation model from the beginning after excluding \(\D_f\) from \(\D\). As indicated by Eq.~\eqref{eq:2}, this is also equivalent to maximizing the posterior distribution over the entire dataset \(\D\) while minimizing the likelihood of the specified interactions \(\D_f\).
We can approximate the posterior $\log P(\varphi | \D_r)$ by leveraging $\varphi_0$ and assuming the prior distribution $g(\varphi)$ as a normal distribution $N(\varphi, \sigma^2)$ as
\begin{equation}
    \mathcal{L} \approx \mathcal{L}_\text{BPR} + (\varphi - \varphi_0)^T \frac{\partial \mathcal{L}_\text{BPR}}{\partial \varphi_0} + \frac{1}{2} (\varphi - \varphi_0)^T \frac{\partial^2 \mathcal{L}_\text{BPR}}{\partial \varphi_0^2} (\varphi - \varphi_0) \label{eq:4}
\end{equation}
At the optimal point, $\mathcal{L}_\text{BPR} \approx 0$ and $\left\| \frac{\partial \mathcal{L}_\text{BPR}}{\partial \varphi_0} \right\|^2 \approx 0$. Then we can derive the approximation of optimal posterior distribution as
\begin{equation}
    \mathcal{L} \approx \frac{1}{2} (\varphi - \varphi_0)^T \frac{\partial^2 \mathcal{L}_\text{BPR}}{\partial \varphi_0^2} (\varphi - \varphi_0) \label{eq:5Sup}
\end{equation}
Based on this, we conclude that maximizing the posterior distribution $\log P(\varphi | \D_r)$ is equivalent to the Reverse Bayesian Personalized Ranking objective in \name, which is to minimize the RPR objective with a regularizer.
\subsection{Time Complexity}
\label{sec:timeComplexity}
The time complexity of unlearning LightGCN with BPR loss can be analyzed based on the dominant operations performed during each training iteration. Here's a breakdown:
\begin{enumerate}
    \item Forward Pass: LightGCN utilizes message passing to propagate information between connected nodes in the graph. This involves operations like matrix multiplication between the embedding matrices and the adjacency matrix. Complexity: $\mathcal{O}(d \cdot |\E| \cdot d_m)$ for each modality. 
    Since there can be multiple modalities, this is summed over all modalities $\mathcal{M}$.
    
    \item Auxiliary Contrastive Loss Calculation: The complexity of the auxiliary contrastive loss calculation depends on the following operations. Refer to \textsc{Mgcn} paper~\cite{yu2023multi} for details.
    \begin{enumerate}
        \item Calculation of Modality Preferences ($P_m$): This step involves a matrix multiplication operation followed by applying a sigmoid non-linearity. Both have a complexity: $\mathcal{O}(d \times d)$. 

        \item Calculation of Modality-shared Features ($E_s$): This step includes attention mechanisms to calculate attention weights ($\alpha_m$) for each modality's features and then obtaining the modality-shared features by aggregating the weighted sum of modality features. 
        Both have a complexity: $\mathcal{O}(d \times |\mathcal{M}|)$. 

        \item Calculation of Modality-specific Features ($\tilde{E}_m$): This step involves subtracting the modality-shared features from the original modality features. Complexity: $\mathcal{O}(d)$.

        \item Adaptive Fusion of Modality-specific and Modality-shared Features ($E_\text{mul}$): This step includes element-wise multiplication ($\odot $) and addition operations, both of which have a complexity: $\mathcal{O}(d)$.

        \item Self-supervised Auxiliary Task Loss ($\mathcal{L}_C$): The complexity of this loss function depends on the number of users ($|\mathcal{U}|$) and items ($|\mathcal{I}|$) and is typically dominated by the number of interactions in the dataset.
    \end{enumerate}
    Summing up the complexities of these steps, we get the total time complexity for the auxiliary contrastive loss calculation: $\mathcal{O}(d \times d+d \times |\mathcal{M}|+d + d+ |\mathcal{U}| + |\mathcal{I}|)$.
    Considering dominant terms only, the complexity is $\mathcal{O}(d^2 + |\mathcal{M}| \cdot d)$.
    
    \item BPR Loss Calculation: This step involves comparing the predicted preference scores for user-item pairs and their corresponding negative samples. 
    
    \textbf{Complexity}: $\mathcal{O}(|\mathcal{U}| \cdot |\mathcal{I}| \cdot K)$, where $|\mathcal{U}|$ is the number of users, $|\mathcal{I}|$ is the number of items, and $K$ is the number of negative samples used per user-item pair.
    
    \item Back-propagation: Back-propagation calculates gradients for all the parameters involved in the forward pass. This involves similar operations as the forward pass but with additional computations for gradients.
    The next step of parameter updates involves updating the model parameters based on the calculated gradients using an optimizer like SGD or Adam. Complexity is similar to the forward pass.
\end{enumerate}
Summing the complexities of each step, we get the total time complexity for unlearning LightGCN with BPR loss and auxiliary contrastive loss in \name:
\begin{align*}
T &= \mathcal{O} \left( \mathcal{M} \cdot d \cdot |\E| \cdot d_m  + |\mathcal{U}| \cdot |\mathcal{I}| \cdot K + d^2 + |\mathcal{M}| \cdot d \right)\\
    \text{where:} \\
    T &= \text{Time complexity} \\
    \mathcal{M} &= \text{Number of modalities} \\
    d &= \text{Embedding dimension} \\
    |\E| &= \text{Number of edges in the graph} \\
    d_m &= \text{Feature dimension for modality } m \\
    |\mathcal{U}| &= \text{Number of users} \\
    |\mathcal{I}| &= \text{Number of items} \\
    K &= \text{Number of negative samples per user-item pair}
\end{align*}

\subsection{Datasets and samples}
\label{sec:datasetSamples}
In accordance with references~\cite{wei2019mmgcn}~\cite{zhang2021mining}~\cite{zhou2023bootstrap}, our research involves experimentation on three distinct categories within the popular Amazon dataset~\cite{hou2024bridging}: (a) Baby, (b) Sports and Outdoors, and (c) Clothing, Shoes, and Jewelry, abbreviated as Baby, Sports, and Clothing respectively. The details of these datasets are outlined in Table \ref{tab:datasets}. As per the methodology described in ~\cite{zhou2023bootstrap}, we utilize pre-extracted visual features consisting of $4096$ dimensions and textual features comprising $384$ dimensions, as previously documented in ~\cite{zhou2023mmrec}.
\begin{table}[H]
\centering
\caption{Statistics of datasets}
\label{tab:datasets}
\begin{tabular}{@{}lllll@{}}
\toprule
\textbf{Dataset} & \textbf{\#Users} & \textbf{\#Items} & \textbf{\#Behaviors} & \textbf{Density} \\ \midrule
Baby             & 19,445          & 7,050           & 1,60,792            & $0.117\%$           \\
Sports           & 35,598          & 18,357          & 2,96,337            & $0.045\%$           \\
Clothing         & 39,387          & 23,033          & 2,78,677            & $0.031\%$           \\ \bottomrule
\end{tabular}%
\end{table}
The datasets from Amazon reviews contains various samples of user interactions with different products. It includes both implicit interactions, such as purchases, and explicit interactions, such as ratings. Additionally, the dataset comprises textual reviews, images of products, and detailed descriptions of the items. \par
\textbf{Sample User and her Features.}\\
\emph{User ID:} AGKASBHYZPGTEPO6LWZPVJWB2BVA \\
\emph{Rating:} 4.0 \\
\emph{Title:} Good buy for preschool naps and home use... \\
\emph{Timestamp:} 1471546337000 \\
\emph{Helpful Vote:} 1\\
\emph{Verified Purchase:} True\\
\emph{Review Text:} I bought two of these for my kids for nap time at their preschool. But they also use at home to "camp"...lol. They seem to work pretty well are cute designs and have held up through multiple washings on the gentle cycle and air fluff or lay flat to dry. They still look brand new to me! They are a little stiff at first ( and sometimes after washing) but with use they do become softer and the inside is plenty soft for sleeping- my kiddos have not complaints!\par
\textbf{Sample Item and its Features.}\\
\begin{figure}[htb]
    \centering
    \includegraphics[width=0.35\textwidth]{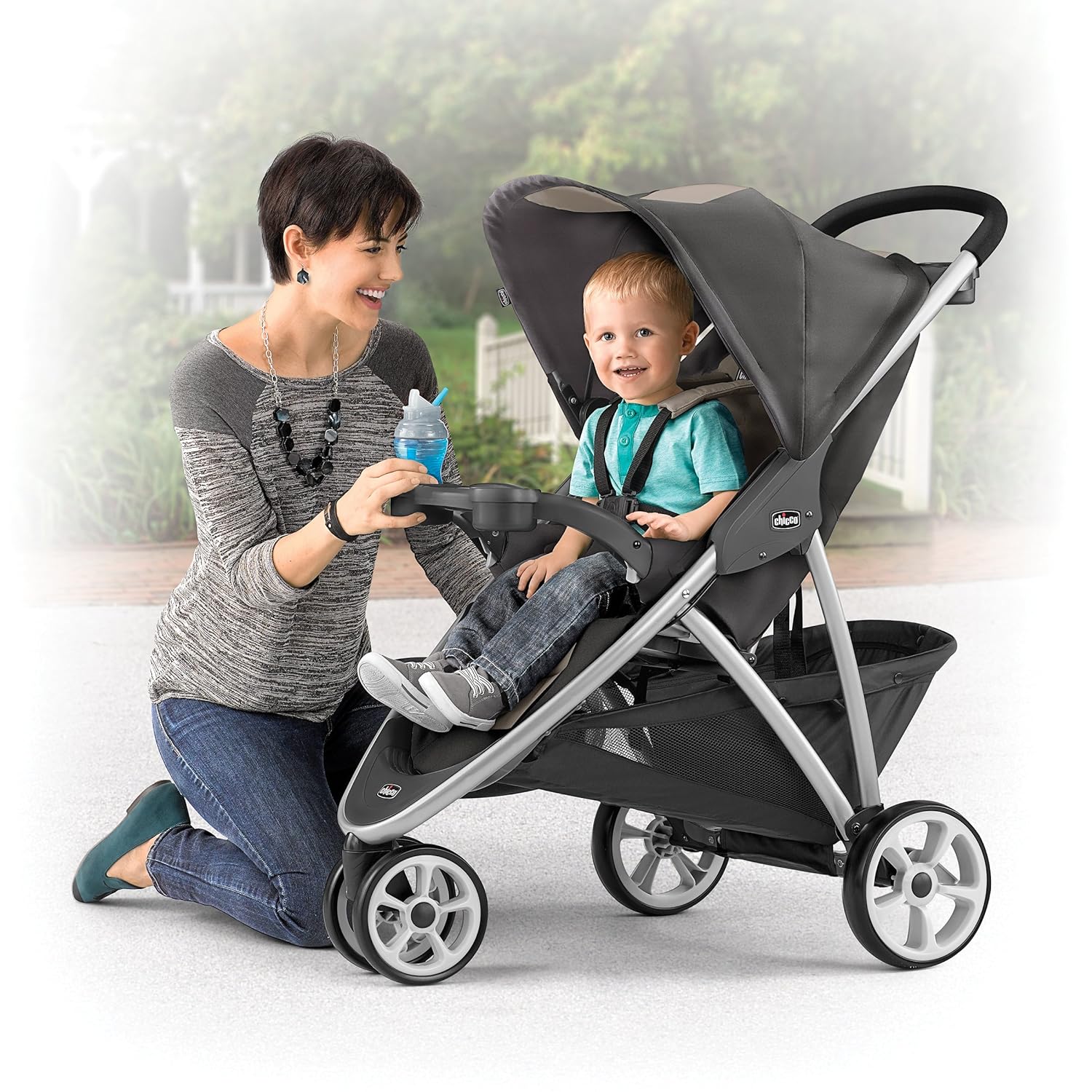}
    \caption{Sample item in the dataset}
\end{figure}
\emph{Title:} Chicco Viaro Travel System, Teak \\
\emph{Average Rating:} 4.6 \\
\emph{Features}: `Aluminum', `Imported', `Convenient one-hand quick fold. Assembled Dimensions- 38 x 25.5 x 41.25 inches. Folded Dimensions- 13.5 x 25.5 x 33.25 inches. Product Weight- 18 pounds', `Ultra-light weight aluminum frame', `3 wheel design allows for nimble steering and a sporty stance', `Front wheel diameter 7 inches and rear wheel diameter 8.75 inches'\\
\emph{Description:} For ultimate convenience, the Chicco Viaro Quick-Fold Stroller has a sleek three-wheel design, lightweight aluminum frame, and one-hand quick fold. A pull-strap and button are conveniently tucked under the seat and easy to activate simultaneously for a compact, free-standing fold. The stroller is even easier to open again after closing. For infants, the Viaro Stroller functions as a travel system with easy click-in attachment for the KeyFit 30 Infant Car Seat. For older riders, the Viaro Stroller includes a detachable tray with two cup holders, adjustable canopy, and multi-position backrest. A swiveling front wheel and suspension help maintain a smooth ride from surface to surface. Toe-tap rear brakes keep the stroller in place when parked. For parents, the Viaro Stroller features a padded push-handle, parent tray with two cup holders, and a large basket that is easily accessible from the front or back.

\subsection{Additional Results}
\label{sec:addResults}
\textbf{Stress testing by unlearning more data.} In our ablation study, we conducted stress testing by systematically unlearning varying percentages of data to evaluate the adaptability of \name to varying unlearning sizes. The results presented in Table \ref{tab:stress} show the model's performance across different subsets of the data, ranging from 5\% to 0.1\%.

\textbf{User Unlearning.} We conduct additional experiments with $K=5, 10$ and $50$. Detailed results for user unlearning  are shown in Table~\ref{tab:userUnlearning2}. 

\textbf{Item Unlearning.} We conduct additional results for item unlearning with NDCG and MAP values and are shown in Table~\ref{tab:itemUnlearning2}.
\begin{table}[H]
\centering
\caption{Recall@20 on validation, test and forget sets when unlearning varying $\%$ users in forget set on the Baby dataset.}
\label{tab:stress}
\resizebox{\columnwidth}{!}{
\begin{tabular}{@{}llllll@{}}
\toprule
Method                                & Set/Percentage & 5\%    & 2.5\%  & 1\%    & 0.1\%  \\ \midrule
\multirow{3}{*}{\textsc{Gold}}        & Valid          & 0.0888 & 0.0926 & 0.0947 & 0.0939 \\
                                      & Test           & 0.0901 & 0.0933 & 0.0951 & 0.0948 \\
                                      & Forget         & 0.0105 & 0.0087 & 0.0099 & 0.0197 \\ \midrule
\multirow{3}{*}{\name} & Valid        & 0.0893 & 0.0887 & 0.0862 & 0.0757 \\
                                      & Test           & 0.0871 & 0.0860 & 0.0837 & 0.0749 \\
                                      & Forget         & 0.0105 & 0.0085 & 0.0099 & 0.0175 \\ \bottomrule 
\end{tabular}
}
\end{table}

\subsection{Discussion}
\label{sec:Discuss}
\textbf{Limitations.}
The method mainly focuses on user-item interactions without considering temporal dynamics, which may be important in some cases. Additionally, it assumes that the data to be forgotten is easily identifiable, which might not hold true in zero-shot unlearning scenarios.

\textbf{Impact.} \name offers significant positive societal impacts, such as enhanced privacy, increased user trust, ethical data handling, and improved recommendation quality by enabling users to control their data and align with privacy regulations like GDPR. However, it also presents potential negative impacts, including the risk of misuse for manipulation, loss of valuable insights, increased computational costs, unintended consequences on model behavior. For example, malicious actors might repeatedly request the unlearning of certain interactions to bias the system in a way that benefits them.

\textbf{Future research} can extend \name to handle more complex types of multi-modal data . Another promising direction is to investigate automated methods for identifying data to be forgotten, improving the practicality and robustness of unlearning in MMRS. 

\end{document}